\documentclass[10pt]{article} %
\usepackage[accepted]{tmlr}

\usepackage{amsmath,amsfonts,bm}

\def\eqref#1{equation~\ref{#1}}

\def\1{\bm{1}}

\DeclareMathAlphabet{\mathsfit}{\encodingdefault}{\sfdefault}{m}{sl}
\SetMathAlphabet{\mathsfit}{bold}{\encodingdefault}{\sfdefault}{bx}{n}

\usepackage[utf8]{inputenc} %
\usepackage[T1]{fontenc}    %
\usepackage{hyperref}       %
\usepackage{url}            %
\usepackage{booktabs}       %
\usepackage{amsfonts}       %
\usepackage{nicefrac}       %
\usepackage{microtype}      %
\usepackage{xcolor}         %
\usepackage{xspace}

\usepackage[pdftex]{graphicx}
\usepackage{amsmath}
\usepackage{amssymb}
\usepackage{bm}
\usepackage{multirow}
\usepackage{pifont}
\usepackage{textcomp}
\usepackage{wrapfig}
\usepackage{lipsum}
\usepackage{subcaption}

\let\cite\citep

\newcommand{\cmark}{\ding{51}}%
\newcommand{\xmark}{\ding{55}}%

\newcommand{\ourscore}{VideoGLUE Score\xspace}
\newcommand{\ourscoreabb}{VGS\xspace}

\newcommand*\samethanks[1][\value{footnote}]{\footnotemark[#1]}

\usepackage[capitalize]{cleveref}
\crefname{section}{Sec.}{Secs.}
\Crefname{section}{Section}{Sections}
\Crefname{table}{Table}{Tables}
\crefname{table}{Tab.}{Tabs.}
\crefname{figure}{Fig.}{Figs.}

\title{VideoGLUE: Video General Understanding Evaluation of Foundation Models}

\author{\vspace{0.5em}
\textbf{Liangzhe Yuan}\thanks{Equal technical contributions.}~\thanks{Corresponding to \texttt{lzyuan@google.com} and \texttt{bgong@google.com}.} \quad \textbf{Nitesh Bharadwaj Gundavarapu}$^{*}$ \quad \textbf{Long Zhao}$^{*}$ \quad \textbf{Hao Zhou}$^{*}$ \quad \\
\vspace{0.5em}
\textbf{Yin Cui}\thanks{Work done at Google. YC is now at NVIDIA; LJ is now at ByteDance; MJ was an intern at Google and is now at Meta.} \quad \textbf{Lu Jiang}\samethanks \quad \textbf{Xuan Yang} \quad \textbf{Menglin Jia}\samethanks \quad \textbf{Tobias Weyand} \quad \textbf{Luke Friedman} \quad \\
\vspace{0.5em}
\textbf{Mikhail Sirotenko} \quad \textbf{Huisheng Wang} \quad \textbf{Florian Schroff}  \quad \textbf{Hartwig Adam} \quad \\
\vspace{0.5em}
\textbf{Ming-Hsuan Yang} \quad \textbf{Ting Liu} \quad \textbf{Boqing Gong}$^\dagger$
\\
\\
Google DeepMind \\
}

\def\month{10}  %
\def\year{2024} %
\def\openreview{\url{https://openreview.net/forum?id=wnI4sJtjqL}} %

\begin{document}

\maketitle

\begin{abstract}
We evaluate the video understanding capabilities of existing foundation models (FMs) using a carefully designed experiment protocol consisting of three hallmark tasks (action recognition, temporal localization, and spatiotemporal localization), eight datasets well received by the  community, and four adaptation methods tailoring an FM for downstream tasks.
Furthermore, we jointly profile FMs'  efficacy and efficiency when adapting to general video understanding tasks using cost measurements during both training and inference.
Our main findings are as follows. 
First, task-specialized models significantly outperform the seven FMs studied in this work, in sharp contrast to what FMs have achieved in natural language and image understanding.  
Second, video-native FMs, whose pretraining data mainly contains the video modality, are generally better than image-native FMs in classifying motion-rich videos, localizing actions in time, and understanding a video of more than one action. 
Third, the video-native FMs can perform well on video tasks under light adaptations to downstream tasks (e.g., freezing the FM backbones), while image-native FMs win in full end-to-end finetuning.
The first two observations reveal the need and tremendous opportunities to conduct research on video-focused FMs, and the last confirms that both tasks and adaptation methods matter when it comes to the evaluation of FMs. 
Our code is released under \url{https://github.com/tensorflow/models/tree/master/official/projects/videoglue}.
 
\end{abstract}

\section{Introduction}
\label{sec:intro}
Foundation model (FM) is a term coined by~\citet{foundation-model}, referring to ``any model that is trained on broad data  that can be adapted (e.g., finetuned) to a wide range of downstream tasks.'' Some representative FMs include but are not limited to BERT~\cite{devlin2018bert}, GPT-3~\cite{brown2020gpt3}, CLIP~\cite{radford2021clip}, and ALIGN~\cite{jia2021align}. 
This work primarily investigates the video understanding capabilies of seven visual and multimodal FMs: CLIP~\cite{radford2021clip}, FLAVA~\cite{singh2022flava}, CoCa~\cite{yu2022coca}, DINOv2~\cite{oquab2023dinov2}, VATT~\cite{akbari2021vatt}, VideoMAE~\cite{tong2022videomae}, and InternVideo~\cite{wang2022internvideo}. 
We select these models because they are amendable for the video understanding and make their checkpoints accessible to us.

It is nontrivial to evaluate FMs. In contrast to ``specialist'' models developed for a particular task, FMs are considered as ``generalists''  that learn shareable meta-knowledge across tasks so that one can quickly adapt them to achieve superior performance on various downstream tasks. Hence, \emph{both the tasks and adaptation methods matter when it comes to the evaluation of FMs.} However, the community has not reached a consensus on these two aspects. FM developers select their own different sets of downstream tasks --- interestingly, often covering no video or only appearance-rich video classification tasks \cite{buch2022cvpr,jie23acl}. Moreover, they rely on distinct adaptation methods, making apples-to-apples comparisons challenging and causing mismatches with  the FMs' actual use cases.

\begin{figure}[t]
\centering
\begin{subfigure}{0.24\linewidth}
\centering
\includegraphics[width=\linewidth]{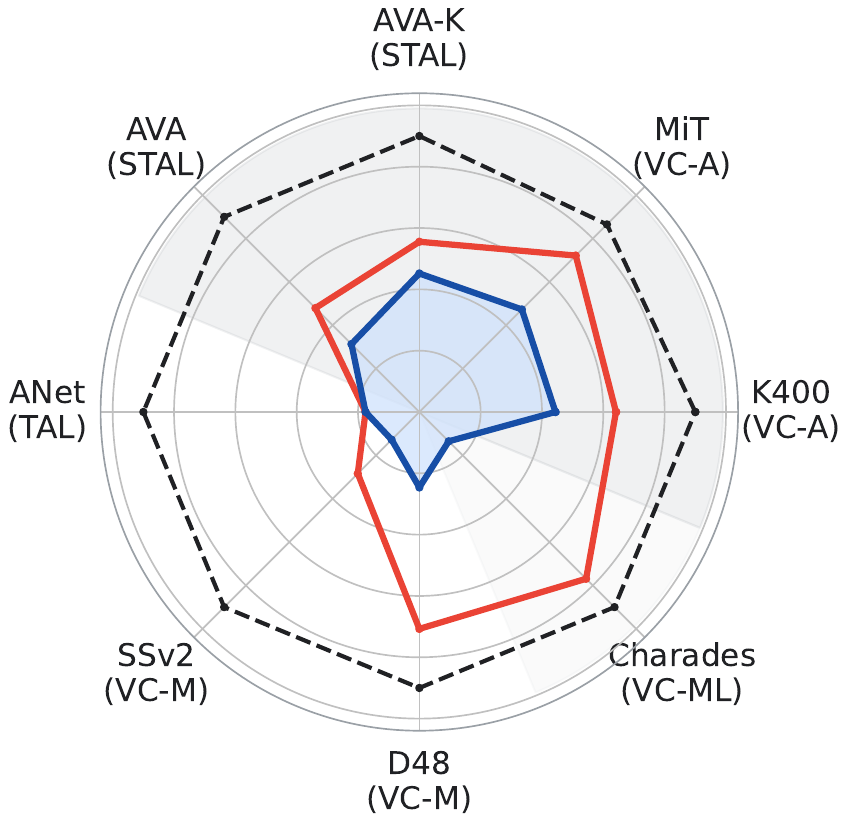}
\caption{CLIP}
\end{subfigure}
\begin{subfigure}{0.24\linewidth}
\centering
\includegraphics[width=\linewidth]{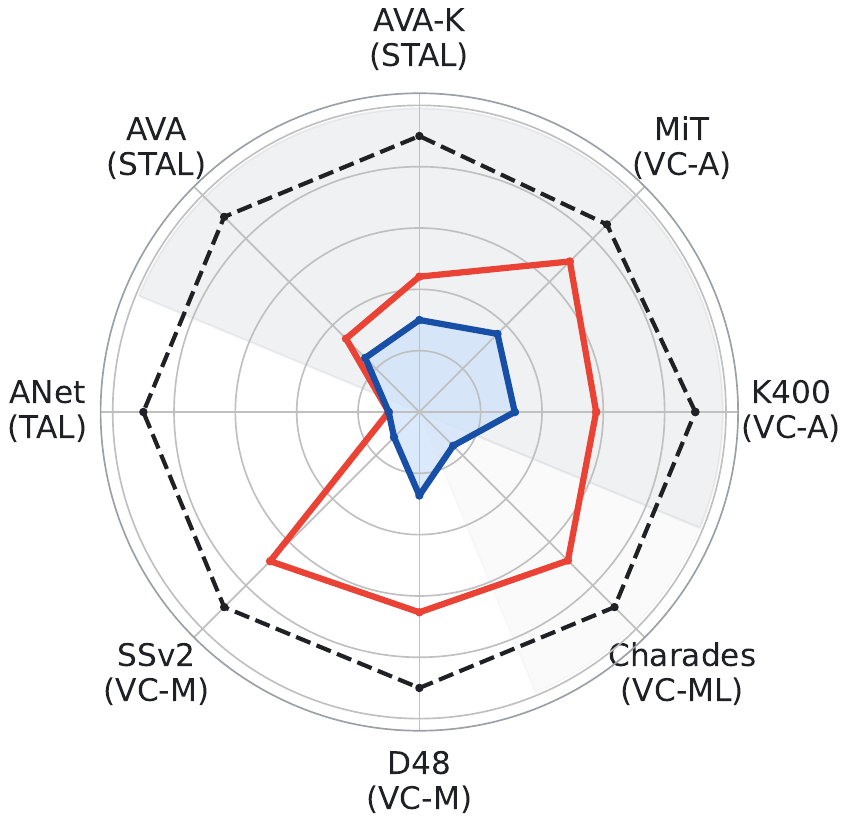}
\caption{FLAVA}
\end{subfigure}
\begin{subfigure}{0.24\linewidth}
\centering
\includegraphics[width=\linewidth]{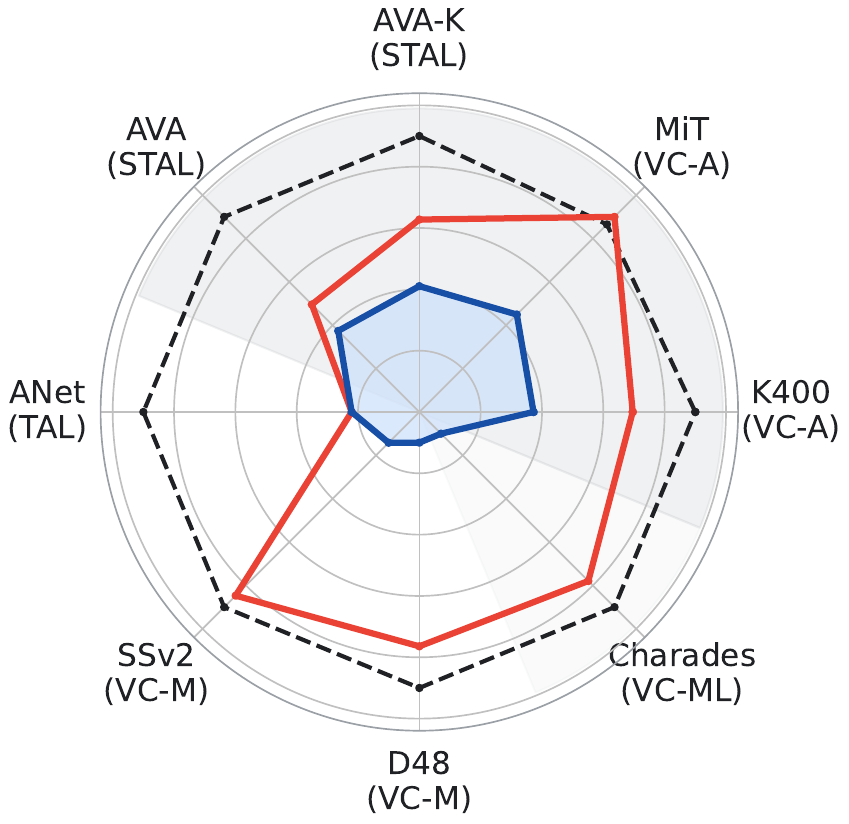}
\caption{CoCa}
\end{subfigure}
\begin{subfigure}{0.24\linewidth}
\centering
\includegraphics[width=\linewidth]{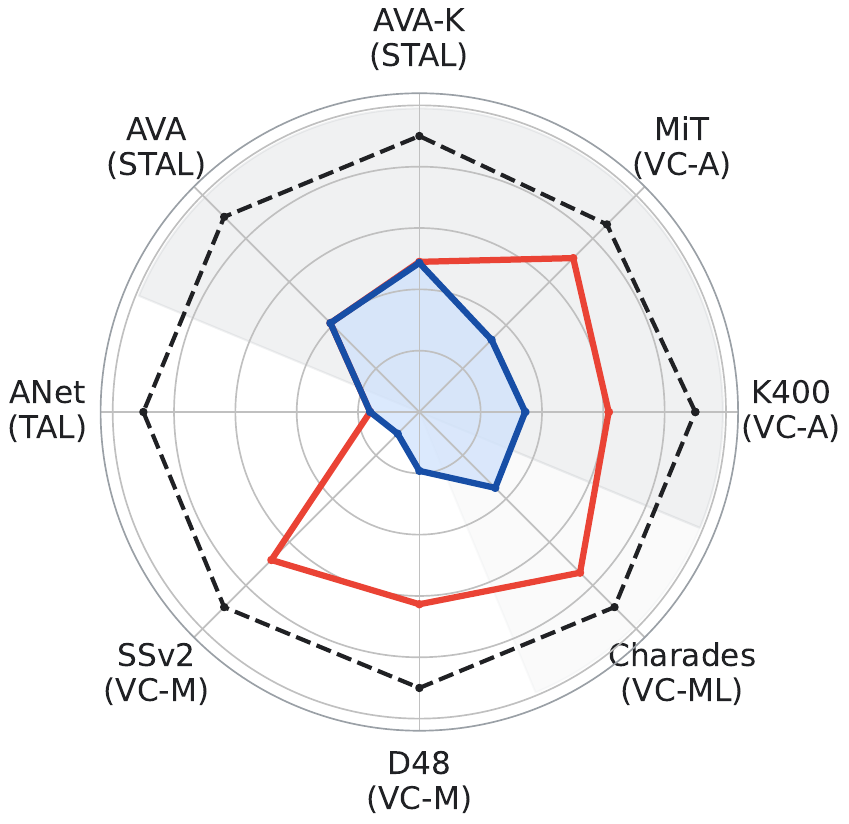}
\caption{DINOv2}
\end{subfigure}
\begin{subfigure}{0.24\linewidth}
\centering
\includegraphics[width=\linewidth]{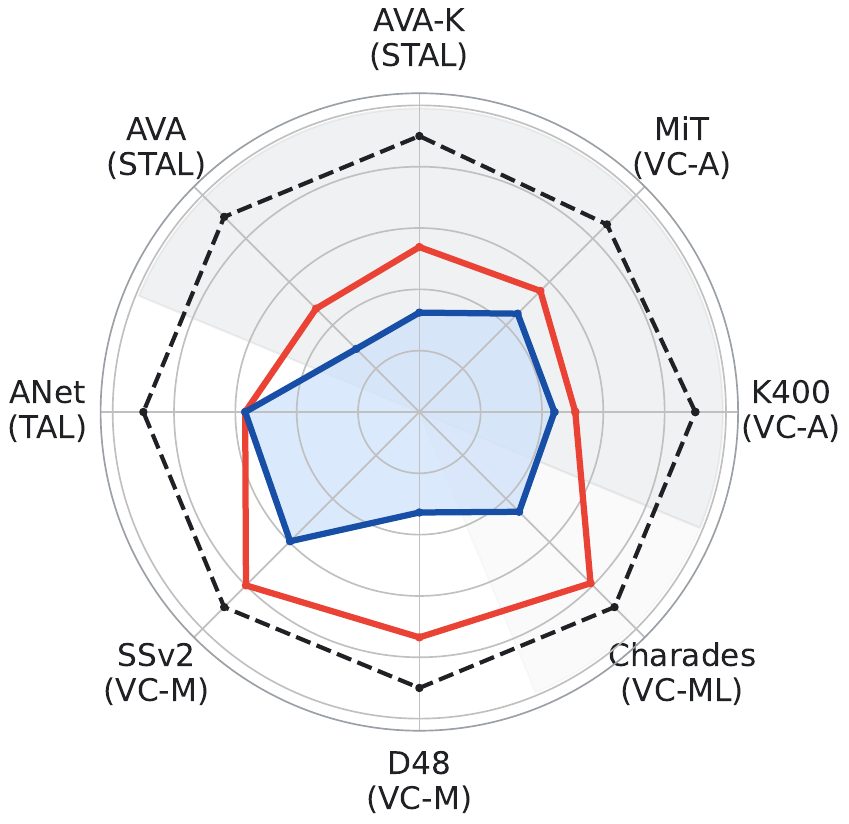}
\caption{VATT}
\end{subfigure}
\begin{subfigure}{0.24\linewidth}
\centering
\includegraphics[width=\linewidth]{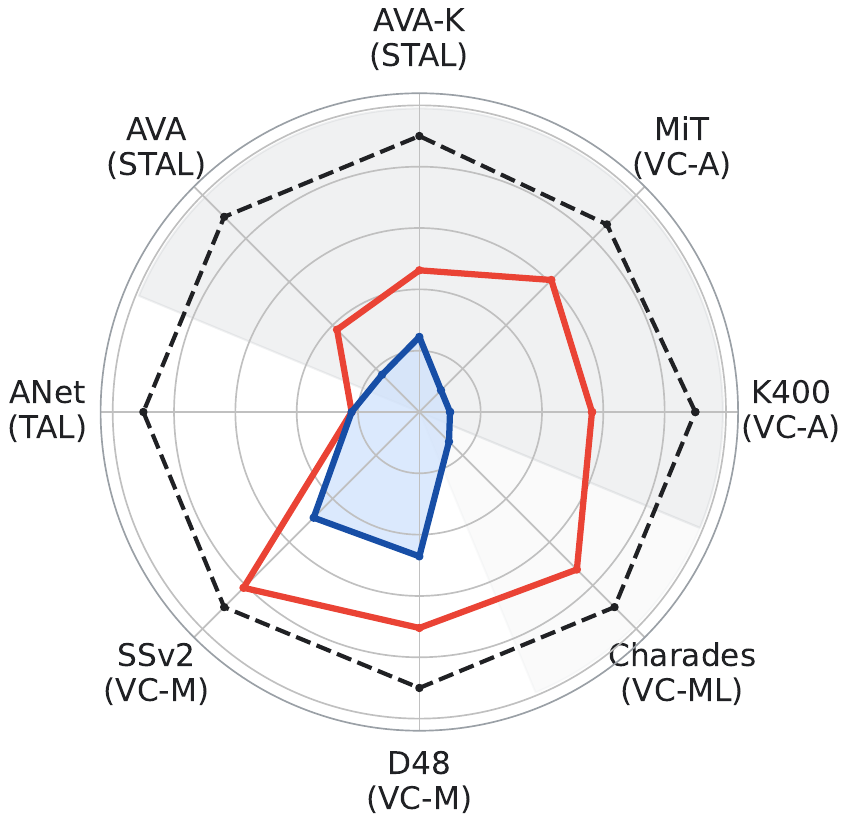}
\caption{VideoMAE}
\end{subfigure}
\begin{subfigure}{0.24\linewidth}
\centering
\includegraphics[width=\linewidth]{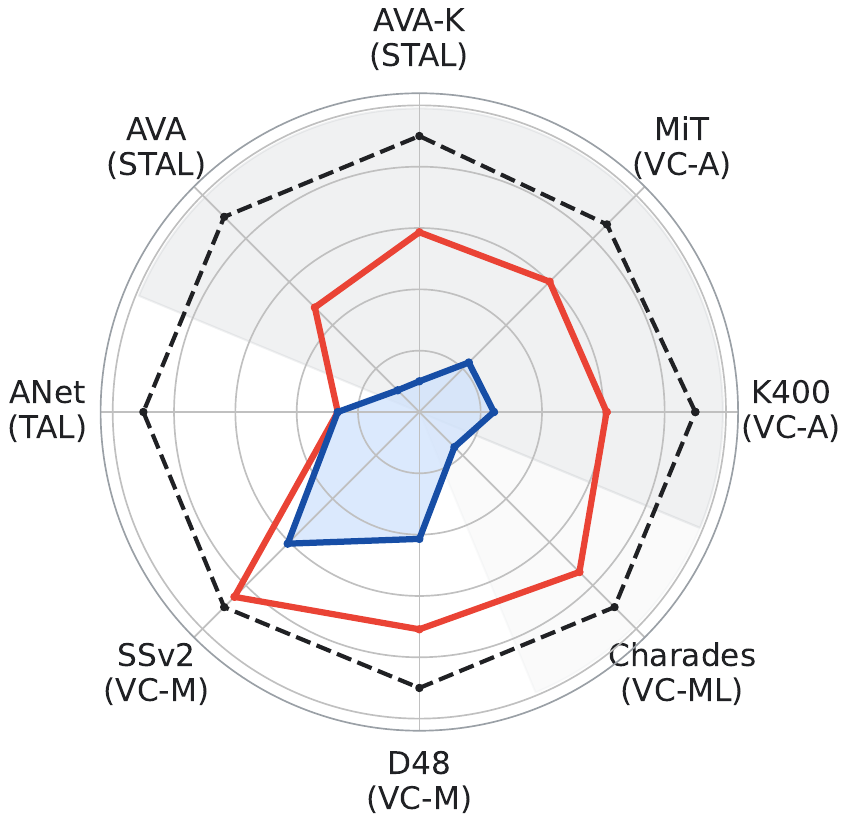}
\caption{InternVideo}
\end{subfigure}

\caption{
Performance of FMs with end-to-end finetuning (red) and frozen backbone (blue), in comparison with state-of-the-art task-specialized models (black) on VideoGLUE benchmarks. VC-A, VC-M, and VC-ML stand for appearance-focused, motion-focused, and multi-label Video Classification tasks, respectively; TAL stands for Temporal Action Localization; STAL stands for Spatiotemporal Action Localization. The highest and lowest performance numbers on each dataset are mapped to $0.9$ and $0.1$, and the other numbers are linearly scaled accordingly on the radar chart. We also use gray shades to represent tasks that are more focused on appearance understanding more than motion. We observe that: (1) FMs generally fall behind task-specialized models; (2) FMs that are trained with video data are generally better than image-native FMs on motion-focused tasks under the frozen backbone setting, and image-native FMs can generally catch up when finetuned end-to-end on the target dataset.
}\label{fig:radar_chart}
\end{figure}

To this end, we propose to evaluate FMs' video understanding capabilities  using a carefully designed experiment protocol, named VideoGLUE, consisting of three hallmark tasks (action recognition, temporal localization, and spatiotemporal localization), eight datasets well received by the research community, and four model adaptation methods tailoring a foundation model for downstream tasks. The tasks examine an FM from various aspects needed for understanding video. The ``all-around'' adaptations represent the main use cases of FMs in the literature and, more importantly, allow us to thoroughly probe an FM's potential in video understanding. 

Why do we specifically focus on videos? The main motivation is to promote video understanding in the evaluation of FMs. More concretely, we test the following conjectures through this work. First, FMs' high performance on existing evaluation suites does not necessarily indicate their potential in video since these suites either lack video-specific tasks or selectively choose video tasks whose appearance feature is more important than motion --- InternVideo~\cite{wang2022internvideo} is an exception as discussed in the next paragraph. 
Second, many existing FMs are unlikely to be able to handle motion in video well, given that they learn primarily from static images~\cite{radford2021clip, singh2022flava, yu2022coca} or short video clips containing limited motion~\cite{feichtenhofer2022vmae, wang2022internvideo}.
Third, popular adaptation methods (e.g., finetuning all weights) cannot supplement FMs with all the cues needed to recognize motion-rich actions and localize entities temporally and/or spatiotemporally, as elaborated in Sections~\ref{sec:e2e_ft} and~\ref{sec:frozen_fms}.

While our work is not the first to emphasize the evaluation of FMs, it is unique on multiple fronts. Unlike ELEVATER~\cite{benchmark-microsoft}'s target of evaluating language-augmented FMs, we consider all FMs adaptable to video understanding which does not necessarily involve language. Unlike Perception Test~\cite{benchmark-deepmind}'s coverage of a broad spectrum of perception tasks, we focus on video, allowing us to cover various aspects of this vertical domain. Interestingly, many of our datasets also appear in InternVideo~\cite{wang2022internvideo}, a video-oriented FM. However, we promote model adaptation methods as an inherent part of the evaluation protocol --- a consistent set of diverse adaptation methods is necessary to provide FMs ample opportunities to expose their video understanding capabilities. Moreover, unlike InternVideo's focus on their single FM, we evaluate FMs developed by different research groups in an uniform experiment protocol --- the first of its kind for visual and multimodal FMs, to the best of our knowledge.

Our main findings are as follows. First, task-specialized models still significantly outperform the seven FMs studied in this work (see Figure~\ref{fig:radar_chart}), in sharp contrast to what FMs have achieved in natural language~\cite{gpt4, roberts2022t5x} and image understanding~\cite{radford2021clip, yu2022coca, chen2022pali}. Hence, there is a need and tremendous opportunities to research video-focused FMs. Second, video-native FMs, whose pretraining data mainly contains the video modality, are generally better than image-native FMs in classifying motion-rich videos, localizing actions in time, and understanding a video of more than one action. Third, the video-native FMs can perform well on video tasks under light adaptations to downstream tasks (e.g., freezing the FM backbones), while image-native FMs win in full end-to-end finetuning. This observation confirms that both tasks and adaptation methods matter when it comes to the evaluation of FMs. 

\section{Related work}
\label{sec:related}

\paragraph{Foundation models.}
One common type of FMs are Large Language Models (LLMs) trained to acquire generic, transferable, and diverse representations that can enable sample-efficient learning and knowledge transfer across a broad range of downstream tasks. 
FMs are often trained with simple self-supervised learning objectives such as predicting the next token in a sentence (e.g., GPT-3~\cite{brown2020gpt3}, PaLM~\cite{chowdhery2022palm}), or denoising the masked tokens (e.g., BERT~\cite{devlin2018bert}, UniLM~\cite{dong2019unified}, and BEiT~\cite{bao2021beit}). 
An intriguing characteristic of FMs is their ability to gradually acquire new capabilities as the model grows and the training data size increases, despite being trained on simple learning objectives~\cite{wei2022emergent}. For example, PaLM~\cite{chowdhery2022palm, anil2023palm}, a massive LM with $540$ billion parameters, has started to show new capabilities in tasks such as explaining jokes, solving math, and performing common-sense reasoning when scaled to over 100B parameters.

In addition to self-supervised transformers, FMs in computer vision also encompass transformers specifically trained to align image-text paired data. These FMs use learning objectives include contrastive learning (e.g., CLIP~\cite{radford2021clip}), denoising masked tokens (e.g., BEiT-3~\cite{wang2022image}), predicting the next token in a single modality (e.g., DALL-E~\cite{ramesh2021zero}) or in the interleaved image-text sequence (e.g., Flamingo~\cite{alayrac2022flamingo}, Kosmos-1~\cite{huang2023language}). Recent FMs are also trained on a mixture of these objectives (e.g., MAE~\cite{he2022masked}, FLAVA~\cite{singh2022flava}, and CoCa~\cite{yu2022coca}).
Our criteria of choosing foundation models to study are primarily based on the definition of FMs, and their amendability on video understanding and accessibility of checkpoints. We leave some models (e.g., detection, segmentation models) out of the scope of this work, because of their current lack of generalization on video understanding tasks. 
Finally we choose seven representative FMs, i.e., CLIP~\cite{radford2021clip}, FLAVA~\cite{singh2022flava}, CoCa~\cite{yu2022coca}, DINOv2~\cite{oquab2023dinov2}, VATT~\cite{akbari2021vatt}, VideoMAE~\cite{tong2022videomae}, and InternVideo~\cite{wang2022internvideo}.

\paragraph{Evaluation of foundation models.}
As the mission of FMs is to enable sample-efficient knowledge transfer, the design of downstream tasks is critical to evaluate the capabilities and limitations of these models. The evaluation of FMs is pioneered by the NLP researchers. For example, GLUE~\cite{benchmark-glue} and SuperGLUE~\cite{benchmark-superglue} introduced a suite of tools for evaluating language understanding tasks.
The authors utilized established public benchmarks and provided tools for evaluating, probing, and benchmarking pretrained FMs, allowing for a comparison to human baselines.
ELEVATER~\cite{benchmark-microsoft} introduced this concept to vision FMs along with a toolkit for evaluating vision-language tasks, including knowledge augmentation, hyperparameter tuning, and three adaptation techniques.
In parallel, there have been attempts to establish a diagnostic benchmark for perceptual understanding of the world. For instance, Perception Test~\cite{benchmark-deepmind} crowd-sourced 11K videos in which about 100 users performed scripted activities. This benchmark comprises videos filmed by only about 100 participants, which may not provide the same level of domain coverage and diversity as the other FM evaluation works mentioned earlier.

\paragraph{Evaluation of video foundation models.}
While some vision-language FMs have incorporated video tasks, their evaluation typically follows that of static images and neglects the unique aspects of video spatial-temporal modeling and reasoning. To our knowledge, no previous work has been solely dedicated to evaluating video FMs. The closest work to ours are InternVideo~\cite{wang2022internvideo} and VideoMAE~\cite{tong2022videomae}, which introduce new FMs and show their superiority over several video datasets. This work has two key differences to the prior ones. First, our evaluation is video-centric using the tasks that require motion understanding or long-term temporal reasoning. Second, instead of promoting new video FMs, our work proposes no new models and is solely dedicated to evaluating current and future video FMs in an impartial reproducible experimental setup. Concretely, our goal is to provide tools for probing and benchmarking FMs on motion tasks in various settings.

\begin{table}
\caption{Foundation models studied in this work (MxM stands for Masked Image/Language/Video Modeling).}
\label{tab:foundation_model}
\begin{center}
\resizebox{0.88\linewidth}{!}{
\begin{tabular}{lcll} 
\toprule
Foundation Model & Pretraining Modality & Pretraining Data & Pretraining Objective\\
\midrule
CoCa~\cite{yu2022coca} & Image + Text & JFT3B + ALIGN & Contrastive + Captioning \\
CLIP~\cite{radford2021clip} & Image + Text & WebImageText & Contrastive \\
FLAVA~\cite{singh2022flava} & Image + Text & PMD & Contrastive + MIM + MLM \\
DINOv2~\cite{oquab2023dinov2} & Image & LVD-142M & MIM + DINO \\
\midrule
VideoMAE~\cite{tong2022videomae} & Video & K400 & MVM \\
InternVideo~\cite{wang2022internvideo} & Video & UnlabeledHybrid & MVM + Contrastive \\
VATT~\cite{akbari2021vatt} & Video + Audio + Text & HT100M & Contrastive \\

\bottomrule
\end{tabular}
}
\end{center}
\end{table}

\section{Tasks and adaptation methods both matter when evaluating foundation models} 
\label{sec:method}
This section describes our video general understanding evaluation (VideoGLUE) benchmark.
We first introduce the visual and multimodal FMs evaluated in this work.
Then we discuss the video-focused downstream tasks and methods to adapt an FM to the tasks. 
The former concretizes the video understanding capabilities we want to evaluate from an FM, while the latter provides various paths for an FM to showcase the corresponding capabilities. 

\subsection{Foundation models for video understanding}\label{sec:foundatin_models} 
We are interested in examining which FMs are good at solving video tasks, what makes them better than others in the video domain, and how to best adapt them to video understanding. 
Table~\ref{tab:foundation_model} shows the seven FMs we gained access to via public repositories or personal communications.
Thanks to the powerfulness and scalability of the transformer architecture~\cite{vaswani2017attention}, most developed FMs converge to adopt the vision transformer architecture. 
Thus for all evaluated FMs, we intentionally choose the ViT-B~\cite{dosovitskiy2020vit} variant to bring fair comparison into our benchmark. 
We also notice, in previous literature, models may be evaluated with different number of frames and resolutions, resulting in unfair comparison~\cite{yan2022videococa, feichtenhofer2021large}. 
In VideoGLUE, we control the number of tokens observed by the ViT to be consistent for different FMs on the same task.
For the detailed setup on each dataset, please refer to supplementary materials~\ref{app:hyperparameters}.

\begin{table}[t!]
\caption{
Summary of statistics, video properties, and data sources of each dataset.
Tasks involved are video classification (VC), spatiotemporal action localization (STAL), and temporal action localization (TAL).
}
\label{tab:datasets}
\begin{center}
\resizebox{0.95\linewidth}{!}{
\begin{tabular}{llcccl} 
\toprule
Task & Dataset & \# of videos (train/validation) & Avg.\ length & Source & Notes \\
\midrule
\multirow{5}{*}{VC} & Kinetics-400 & $235,693$ / $19,165$ & $10$ secs & Web & Holistic, appearance \\
 & Moments in Time & $791,246$ / $33,898$ & $3$ secs & Web & Holistic, appearance \\
 & Something-Something v2 & $168,913$ / $24,777$ & $2\sim 6$ secs & Crowdsource & Holistic, motion \\
 & Diving48 & $15,027$ / $1,970$ & $5$ secs & Web & Holistic, motion \\
 & Charades & $7,811$ / $1,814$ & $30$ secs & Crowdsource & Multi-label, long-clip \\
\midrule
TAL & ActivityNet v1.3 & $10,002$ / $4,926$ & $5\sim 10$ mins & Web & Temporal \\
\midrule
\multirow{2}{*}{STAL} & AVA v2.2 & $210,634$ / $57,371$ & $15$ mins & Movie & Spatiotemporal, instance \\
 & AVA-Kinetics & $354,201$ / $91,919$ & $10$ secs & Web & Spatiotemporal, instance \\
\bottomrule
\end{tabular}}
\end{center}
\end{table}

\subsection{Video understanding tasks} \label{sec:video-tasks}

Like objects' role in image understanding, actions are the core of video understanding, leading us to select tasks and datasets that \emph{recognize} and \emph{localize} actions in time and space. Table~\ref{tab:datasets} provides a quick summary. Next, we explain the rationale behind the particular choices of datasets and postpone the datasets' details to the supplementary materials~\ref{app:datasets}.

\subsubsection{Recognizing actions}
\noindent\textbf{General actions.} 
We first include the action recognition datasets of Kinetics-400 (K400)~\cite{kay2017kinetics}, Moments in Time (MiT)~\cite{mit}, and Charades~\cite{sigurdsson2016charades}, considering their popularity and they are complementary to each other. 
Data domain coverage is an important factor when designing benchmarks for FMs, as nowadays FMs are typically trained on massive data sources.
K400 videos are from YouTube, MiT draws videos from different Web venues, while Charades contains scripted indoor videos. 
Internet often returns entertaining and atypical videos, while Charades is about typical everyday videos~\cite{sigurdsson2016charades}.
Regarding action labels, the datasets differ in granularities and real-life scenarios: a verb defines an action in MiT, K400 groups actions by verb-subject pairs, and Charades actions are about indoor activities. 
Regarding the average length, K400 and MiT videos are between 3 and 10 seconds, each with one action label, while Charades videos are about 30 seconds, each with multiple actions.

\noindent\textbf{Fine-grained motion-focused actions.} We also include Something-something v2 (SSv2)~\cite{goyal2017something} and Diving48 (D48)~\cite{li2018diving48} as another two action recognition datasets, whose actions are fine-grained and motion-focused. SSv2 contains 174 human hand gestures as action labels, such as putting something into something, turning something upside down, and covering something with something.
The videos are mostly focusing on hand-object interactions.
D48 videos are all about competitive diving recordings collected from Web sources.
Notably, in these datasets the foreground objects' motion is a more significant discriminative cue than their appearance.

\subsubsection{Localizing actions}
The videos in action recognition are trimmed, but actions could occur anywhere in a video in the wild. Hence, temporal and spatiotemporal action localization is also crucial to video understanding. Accordingly, we choose three datasets for the experiments: the action localization track of ActivityNet v1.3 (ANet)~\cite{caba2015activitynet}, Atomic Visual Actions (AVA)~\cite{gu2018ava}, and AVA-Kinetics (AVA-K)~\cite{li2020avak}. The last two require a model to localize and recognize actions in both time and space, and their underlying videos are movies and general YouTube videos, respectively.

\subsection{Adaptation methods}

In this section, we detail the task-specific neural architecture design and adaptation methods when applying FMs to downstream tasks.

\subsubsection{Modifying foundation model architectures for downstream tasks}
Given an $\textsc{fm}(\cdot)$, we can apply $\textsc{fm}(\cdot)$ to a video clip $C$ to extract a set of $k$ feature maps $\{F\}^{k}=\textsc{fm}(C), F \in \mathbb{R}^{n \times h \times w \times c}$, where $k$ is the number of endpoint layers from an FM, and $n, h, w, c$ are respectively a feature map's length, height, width, and number of channels. 

For video classification tasks, we cast a feature map $F$ as $n \times h \times w$ tokens and aggregate them into a global representation using a learnable query token $\tau$ and lightweight cross-attention layers~\cite{dosovitskiy2020vit}. %
For spatiotemporal action localization, following the standard practice~\cite{feichtenhofer2019slowfast, tong2022videomae}, we first detect humans on key-frames using a human detector~\cite{ren2015faster}, producing a set of human bounding boxes $B$.
We then apply the RoI pooling operation~\cite{jaderberg2015spatial_transformer} that takes both the feature map $F$ and box coordinates $B$ as inputs and outputs one feature vector per box as the query token, $\tau = \textsc{RoIPool}(F, B)$, followed by the same cross-attention layers as in video classification. For both groups of tasks, we stack a linear classifier on top of the task token's last-layer encoding for final classification:
\begin{equation}
    p = \textsc{LinearClassifier}(\textsc{CrossAttention}(\tau, F)).
\end{equation}

For temporal action localization, we first perform feature extraction in a sliding window manner, resulting in a sequence of globally average pooled features $\{\textsc{AvgPool}(F_1), \cdots, \textsc{AvgPool}(F_t)\}$ for each video. 
Following a popular choice of prior works~\cite{alwassel2021tsp, ju2022prompting, liu2022empirical}, we employ G-TAD~\cite{xu2019gtad} as our task head for predicting the action category and its start and end timestamps.

\begin{figure}[t]
\centering
\begin{subfigure}{0.195\linewidth}
\centering
\includegraphics[height=1.8\linewidth]{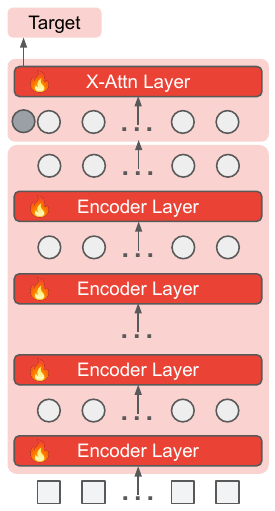}
\caption{}
\end{subfigure}
\begin{subfigure}{0.195\linewidth}
\centering
\includegraphics[height=1.8\linewidth]{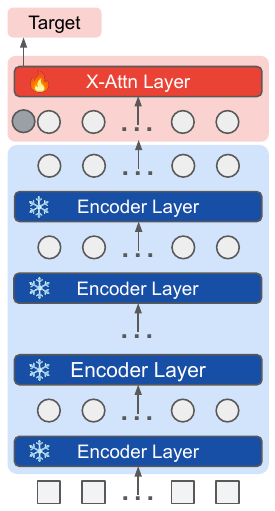}
\caption{}
\end{subfigure}
\begin{subfigure}{0.39\linewidth}
\centering
\includegraphics[height=0.9\linewidth]{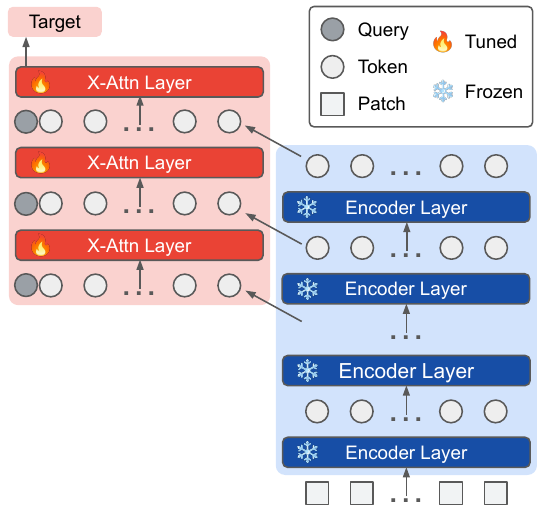}
\caption{}
\end{subfigure}
\begin{subfigure}{0.195\linewidth}
\centering
\includegraphics[height=1.8\linewidth]{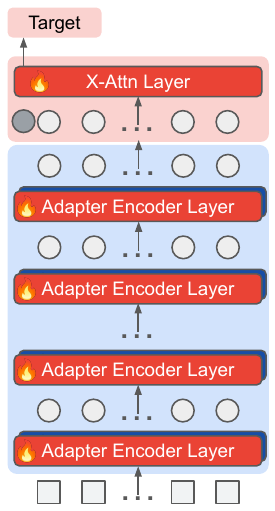}
\caption{}
\end{subfigure}

\caption{
We study four adaptation methods to apply a foundation model (FM) to video understanding downstream tasks: (a) end-to-end finetuning, (b) frozen backbone, (c) frozen backbone with multi-layer attention pooler (MLAP), and (d) a low-rank adapter.
}\label{fig:model_adaptations}
\end{figure}

\subsubsection{Adapting modified foundation model to downstream tasks}
Adapting the modified FMs to a downstream task is to tune their weights. Then, we immediately have two basic adaptation strategies: (1) full finetuning to update all weights in the original FM plus the task head and (2) freezing FM weights and only updating newly added weights. {The choice of the adaptation methods depends on specific application scenarios such as computation and memory constraints. We argue that an ideal FM should perform well across various adaptation methods to support the breadth of use cases.}

\noindent \textbf{End-to-end finetuning.} 
{End-to-end finetuning is the most common FM evaluation method for videos \cite{akbari2021vatt,feichtenhofer2022vmae,tong2022videomae,wang2022internvideo}, but it requires the deployment of a separate and possibly expensive FM for each downstream task.} When finetuning all weights in the modified FMs, we limit cross-attention to a single  transformer layer with 12 heads and hidden size 768. 
We vary learning rates and weight decays for each experiment to ensure every FM is configured to its best setup. Figure~\ref{fig:model_adaptations}(a) illustrates this end-to-end finetuning. 

\noindent{\textbf{Freezing foundation model weights.}} Linear probing and cross-attention based pooling over frozen FM features are routinely used to test the strength of the FM representation \cite{tong2022videomae,yu2022coca,singh2022flava,he2022masked,lin2022frozen}. In practice, adapting task-specific heads with a frozen FM allows us to deploy the same FM for multiple tasks. If we use light-weight heads over the FM features, then a single FM inference can serve multiple tasks efficiently in terms of both compute and memory. To this end, we examine two variations with a frozen FM, one with a single cross-attention layer and the other with multiple layers. The first results in exactly the same model architectures as in end-to-end finetuning (Figure~\ref{fig:model_adaptations}(b)), and the second allows us to leverage an FM's hierarchical features beyond its last endpoint layer (Figure~\ref{fig:model_adaptations}(c)).  
First, the frozen features are extracted from the last $k$ layers, $F_{N-k+1}$, $F_{N-k+2}$, ..., $F_N$. Then, attention pooling is applied between a learnable token $\tau$ and the features $F_{N-k+1}$ using multi-head cross-attention (MHCA). The output of this layer serves as the query token for the next round of attention pooling with the features $F_{N-k+2}$.
This process is repeated for $k$ rounds:
\begin{equation}
\begin{aligned}
\tau_{N-k+1} &= \textsc{MLP}(\textsc{MHCA}(\tau, F_{N-k+1})), \\
\tau_{N-k+2} &= \textsc{MLP}(\textsc{MHCA}(\tau_{N-k+1}, F_{N-k+2})), \\
&... \\
\tau_{N} &= \textsc{MLP}(\textsc{MHCA}(\tau_{N-1}, F_{N})),
\end{aligned}
\end{equation}
where $k=4$ in our experiments, and the final classifier is $p = \textsc{LinearClassifier}(\tau_N)$.

\noindent{\textbf{Freezing foundation model weights with low-rank adaptation}}. 
Finally, we explore a frozen FM beyond the last $k$ layers using a low-rank adapter~\cite{hu2021lora}, which is a bottleneck architecture that projects a feature tensor into a low-dimensional space and then up-samples to the original space. The bottleneck space's dimension is 64 in our experiments. Through inserting a few adapter layers with trainable weights $\{w\}$ into the pretrained FM while keeping all FM's weights frozen, the feature adapter is more parameter-efficient than end-to-end finetuning the whole network while achieving better performance than simply adding a task head to the frozen FM. 
Essentially, the adapter leads to a new FM with some trainable weights $\{w\}$:
    $\tilde{F} = \widetilde{\textsc{FM}}(C, \{w\}),$
such that the output feature maps remain the same in shape as the original FM's output (Figure~\ref{fig:model_adaptations}(d)). Hence, different pooling schemes and task heads aforementioned could be applied to the extracted feature map $\tilde{F}$. For simplicity, we still choose the single-layer cross-attention as the default task head due to its computation efficiency and performance. 

{The low-rank adaptation allows  a single FM for multiple tasks, in contrast to the per-task models in  end-to-end finetuning. However, it incurs a per-task  forward pass at inference time, being less efficient than the task-specific heads over frozen features.}

\section{Experiments}
\label{sec:experiment}

\newcommand{\eat}[1]{}
\eat{%
\subsection{Experimental settings}
FMs performance on video understanding tasks could be affected not only by the model quality itself, but also by the setup of evaluations.
Many recent advances on task-specific modeling could boost FMs performance on downstream tasks, resulting in an un-controlled comparison.
For instance, applying stronger data augmentations like AutoAug~\cite{cubuk2018autoaugment} and CutMix~\cite{yun2019cutmix} would result in improvements on certain datasets, e.g., K400 and SSv2, while they degrade the model performance on D48 and MiT, according to our findings.

Larger image resolution and wider temporal coverage of a video clip tend to improve the task metrics, at a cost of overall higher memory footprint and computation budget.
Taking the widely adopted vision transformer~\cite{dosovitskiy2020vit} as an example, even with the same ViT-b backbone, CoCa uses patch size $18$ to tokenize an image while other methods usually apply patch size $16$. 
If we apply the same image resolution $288\times288$ for all models, CoCa could have much longer token sequence than other methods.
Similarly, people usually pull the same number of video frames to evaluate different FMs. 
However, we notice that VATT~\cite{akbari2021vatt} uses first-layer tokenizer with temporal kernel size 4 while VideoMAE~\cite{tong2022videomae} uses temporal kernel size 2. 
Applying the same number of frames to both models would effectively result in using half-length tokens by VATT, which could result in significant performance drop.
Observations akin to the aforementioned ones motivate us to create a consistent experiment environment and pull FMs of interest together for a thorough and comprehensive study.

The philosophy of our experimental design is to apply the same computational budget for each FM during adaptations.
In our experiment, we only allow limited flexibility on tuning learning rate and weight decay with downstream task adaption while keeping other settings identical across different FMs on each dataset.
This setup could rule-out distracting factors and help us draw insights on feature quality and generalization capability on video understanding of FMs more clearly than ever before.

Table~\ref{tab:e2e_table},~\ref{tab:frozen_table},~\ref{tab:evl_table},~\ref{tab:adapter_table} show the results of evaluating each candidate FM using the four different adaptation methods proposed in this paper.
In the tables, we compute the average scores by using different adaptation methods.
The average score is calculated by first averaging metric $s$ of $G$ datasets in one dataset group (e.g., K400 and MiT under VC-A) and then take the average across $M$ different tasks:
\begin{equation}
    \mathcal{S} = \frac{1}{M} \sum^{i=1}_M \frac{1}{G_i} {\sum^{j=1}_{G_i} s_{i,j}}.
\end{equation}
}%

\subsection{End-to-end finetuning}
\label{sec:e2e_ft}
Table~\ref{tab:e2e_table}  shows the end-to-end finetuning results of six FMs on eight datasets. We split the FMs into two groups based on their input modalities at the time of pretraining: CLIP, FLAVA, CoCa, and DINOv2 are image-native FMs, while VATT, VideoMAE, and InternVideo are video-native. The datasets span  video classification (VC) and spatiotemporal action localization (STAL).
Note that it is infeasible to end-to-end fine-tune or LoRA fine-tune the vision encoder on TAL task, because the videos in ANet are typically long (up to 30 minutes). We follow the common practice of pre-computing visual features in a sliding window manner offline, and training a temporal detection network on top of the visual features~\cite{wang2021proposal, zhang2022actionformer}. We will report TAL results in the next section.
We draw the following observations from Table~\ref{tab:e2e_table}.

\begin{table}[t!]
\caption{
Evaluating FMs when adapted to video understanding tasks using end-to-end finetuning.
We report the Top-1 accuracy on K400, MiT, SSv2 and D48, MAP on Charades and ANet, and mAP@IOU0.5 on AVA and AVA-K.
}
\label{tab:e2e_table}
\begin{center}
\resizebox{0.96\linewidth}{!}{
\begin{tabular}{lcccccccccc}
\toprule
 
 \multirow{2}{*}{Models} & \multicolumn{2}{c}{VC-A)} & & \multicolumn{2}{c}{VC-M} & VC-ML  & TAL & \multicolumn{2}{c}{STAL} & \multirow{2}{*}{Avg.} \\
\cline{2-3} \cline{5-6} \cline{9-10}
& K400 & MiT & & SSv2 & D48 & Charades & ANet & AVA & AVA-K &
\\
\midrule
CLIP & $81.0$ & $39.0$ & & $46.6$ & $75.7$ & $54.3$ & $-$ & $27.1$ & $28.9$ & $52.8$
\\
FLAVA &$79.1$ & $38.3$ & & $61.1$ & $72.0$ & $48.6$ & $-$ & $22.0$ & $25.6$ & $49.4$ 
\\
CoCa & $\textbf{82.6}$ & $\textbf{43.6}$ & & $66.8$ & $\textbf{79.6}$ &  $55.0$ & $-$ & $\textbf{27.7}$ & $\textbf{31.0}$ &  $55.2$ 
\\
DINOv2 & $80.3$ & $38.7$ & & $60.9$ & $70.2$ & $52.4$ & $-$ & $24.6$ & $27.0$ & $50.5$
\\
\midrule
VATT & $77.1$ & $34.8$ & & $65.1$ & $77.6$ &  $\textbf{55.7}$ & $-$ & $27.0$ & $28.4$ & $52.7$ 
\\
VideoMAE & $78.7$ & $36.1$ & & $65.5$ & $75.5$ & $51.4$ & $-$ & $23.5$ & $26.2$ & $51.0$ 
\\
InternVideo &$80.1$ & $35.9$ & & $\textbf{67.0}$ & $75.8$ & $52.2$ & $-$ & $27.2$ & $29.8$ & $52.5$ 
\\

\midrule
Task- & $88.6$ & $42.7$ & & $68.7$ & $88.9$ & $63.2$ & $37.5$ & $42.3$ & $38.9$ & \multirow{2}{*}{$-$} \\
specialized & \footnotesize{(TubeViT)} & \footnotesize{(UniformerV2)} & & \footnotesize{(MViT)} & \footnotesize{(AIM)} & \footnotesize{(MoViNet)} & \footnotesize{(PRN)} & \footnotesize{(RAFT)} & \footnotesize{(RAFT)}
\\
\bottomrule
\end{tabular}}
\end{center}
\end{table}

\textit{FMs underperform task-specialized models on video tasks in general.} 
Table~\ref{tab:e2e_table}'s last row collects the state-of-the-art results on the eight datasets, each obtained by a task-specialized model with comparable architecture or size to ours in the prior work. 
Specifically, those task-specialized models are RAFT~\cite{rajasegaran2023benefits}, PRN~\cite{wang2021proposal}, TubeViT~\cite{piergiovanni2023rethinking}, UniformerV2~\cite{li2022uniformerv2}, AIM~\cite{yang2023aim}, MViT~\cite{fan2021multiscale}, and MoViNet~\cite{kondratyuk2021movinets}, respectively.
All seven FMs underperform the task-specialized models on all video tasks except on Moments in Time at the comparable model scale, indicating the lack of strong video-focused FMs.
This observation is in sharp contrast to what FMs have achieved on natural language~\cite{gpt4, anil2023palm} and image understanding~\cite{chen2022pali}.

\textit{CoCa performs the best among image-native FMs on the video tasks.} It actually gives rise to the highest accuracy on all datasets, with slightly inferior performance on SSv2 and Charades. This shows strong generalization capability of the CoCa model, regardless it is an image-based model with image-only pre-training data.  
However, in the latter session, we will reveal that under light-weight and parameter efficient adaptation scenarios, the same model may perform inferior on many video understanding tasks, especially on SSv2 (Tables~\ref{tab:frozen_table},~\ref{tab:evl_table},~and~\ref{tab:adapter_table}), Charades~(Tables~\ref{tab:frozen_table},~\ref{tab:evl_table},~and~\ref{tab:adapter_table}), and ANet~(Tables~\ref{tab:frozen_table},~and~\ref{tab:evl_table}), which require complex motion or  multiple actions understanding per video.
These are in contrast that CoCa achieves the best general performance in end-to-end fine-tuning (Table~\ref{tab:e2e_table}), highlighting the importance of considering adaptation methods on FMs benchmarking.

\subsection{Freezing foundation models}
\label{sec:frozen_fms}
End-to-end finetuning is infeasible for some application scenarios due to FMs' rapidly growth in size and the consequent demands in computational resources. %
In the following, we evaluate frozen FMs with various adaptation methods. Tables~\ref{tab:frozen_table},~\ref{tab:evl_table},~and~\ref{tab:adapter_table} are the results of adaptation with a single cross-attention layer, multiple cross-attention layers, and a low-rank adapter, respectively. %

\begin{table}[t!]
\caption{
Evaluating FMs when adapted to video understanding using frozen features. Only weights in the task heads are updated using the downstream tasks' training sets.
}
\label{tab:frozen_table}
\begin{center}
\resizebox{0.74\linewidth}{!}{
\begin{tabular}{lcccccccccc} 
\toprule
 
 \multirow{2}{*}{Models} & \multicolumn{2}{c}{VC-A} & & \multicolumn{2}{c}{VC-M} & VC-ML & TAL & \multicolumn{2}{c}{STAL} & \multirow{2}{*}{Avg.} \\
\cline{2-3} \cline{5-6} \cline{9-10}
 & K400 & MiT & & SSv2 & D48 & Charades & ANet & AVA & AVA-K & \\

\midrule
CLIP & $\textbf{75.2}$ & $\textbf{32.6}$ & & $41.0$ & $44.1$ & $11.2$ & $32.7$ & $21.1$ & $25.9$ &  $32.8$ \\
FLAVA & $71.3$ & $29.7$ & & $40.6$ & $45.9$ &$12.6$ &$32.2$ & $18.8$ & $21.5$ & $31.7$ \\
CoCa & $73.1$ & $32.0$ & & $41.5$ & $34.1$  & $8.8$ & $33.0$ & $23.3$ & $24.7$ & $31.2$ \\
DINOv2 & $72.3$ & $29.0$ & & $40.0$ & $40.4$ & $25.8$ & $32.6$ & $\textbf{24.6}$ & $\textbf{26.9}$ & $35.0$ \\
\midrule
VATT &$75.1$ & $32.1$ & & $57.8$ & $49.7$ & $\textbf{33.3}$ & $\textbf{35.3}$ & $20.3$ & $22.2$ & $39.1$ \\
VideoMAE & $65.1$ & $23.0$ & & $53.9$ & $\textbf{59.5}$ & $11.3$ & $33.0$ & $16.0$ & $19.9$ & $32.6$ \\
InternVideo & $69.3$ & $26.3$ & & $\textbf{58.2}$ & $55.6$ & $13.0$ & $33.3$ & $13.4$ & $15.7$ & $33.1$ \\
\bottomrule
\end{tabular}}
\end{center}
\end{table}

\begin{table}[t!]
\caption{
Evaluating FMs when adapted to video understanding using multi-layer attention pooler (MLAP), which takes multiple frozen features from an FM as inputs and map them hierarchically for the final task prediction. Only the multi-layer attention pooling layers are updated using the downstream tasks' training sets. 
}
\label{tab:evl_table}
\begin{center}
\resizebox{0.74\linewidth}{!}{
\begin{tabular}{lcccccccccc}
\toprule
 
 \multirow{2}{*}{Models} & \multicolumn{2}{c}{VC-A} & & \multicolumn{2}{c}{VC-M} & VC-ML & TAL & \multicolumn{2}{c}{STAL} & \multirow{2}{*}{Avg.}\\
\cline{2-3} \cline{5-6} \cline{9-10}
 & K400 & MiT & & SSv2 & D48 & Charades & ANet & AVA & AVA-K & \\

\midrule
CLIP & $\textbf{77.1}$ & $\textbf{39.0}$ & & $50.1$ & $55.8$ & $41.5$ & $33.9$ & $\textbf{27.7}$ & $\textbf{29.6}$ & $43.3$ \\
FLAVA & $71.5$ & $34.5$ & & $43.1$ & $58.5$ & $38.2$ & $32.4$ & $21.3$ & $23.2$ & $39.3$ \\
CoCa & $74.2$ & $37.2$ & & $45.9$ & $48.4$ &$19.6$ & $33.3$ & $24.4$ & $27.0$ & $36.3$ \\
DINOv2 & $75.4$ & $36.0$ & & $46.3$ & $51.9$ & $47.8$ & $33.6$ & $25.4$ & $27.0$ & $42.5$ \\
\midrule
VATT & $75.1$ & $35.6$ & & $58.7$ & $60.1$ & $\textbf{58.2}$ &$\textbf{35.0}$ & $22.9$ & $24.1$ & $46.3$ \\
VideoMAE & $71.7$ & $32.2$ & & $57.4$ & $69.6$ &$35.9$ & $33.4$ & $19.6$ & $22.1$ & $40.9$ \\
InternVideo & $73.7$ & $34.7$ & & $\textbf{60.3}$ & $\textbf{71.9}$ &$40.5$ & $33.6$ & $15.9$ & $17.7$ & $42.2$ \\
\bottomrule
\end{tabular}}
\end{center}
\end{table}

\begin{table}[t!]
\caption{
The low-rank adapter results of FMs for video understanding. 
We only update the weights of the adapter and task head while keeping the original FMs' weights  frozen.
}
\label{tab:adapter_table}
\begin{center}
\resizebox{0.74\linewidth}{!}{
\begin{tabular}{lcccccccccc}
\toprule
 
 \multirow{2}{*}{Models} & \multicolumn{2}{c}{VC-A} & & \multicolumn{2}{c}{VC-M} & VC-ML & TAL &  \multicolumn{2}{c}{STAL} & \multirow{2}{*}{Avg.} \\
\cline{2-3} \cline{5-6} \cline{9-10}
& K400 & MiT & & SSv2 & D48 & Charades & ANet & AVA & AVA-K & \\

\midrule
CLIP & $80.2$ & $39.7$ & & $56.0$ & $\textbf{77.2}$ & $44.2$ & $-$ & $24.5$ & $28.0$ & $49.3$ \\
FLAVA & $74.7$ & $34.1$ & & $52.1$ & $68.4$ & $40.8$ & $-$ &  $17.9$ & $23.8$ & $44.1$ \\
CoCa & $\textbf{80.9}$ & $\textbf{41.4}$ & & $56.1$ & $67.1$ & $45.8$ & $-$ & $\textbf{26.6}$ & $\textbf{28.7}$ & $49.0$ \\
DINOv2 & $77.7$ & $36.1$ & & $59.0$ & $76.6$ & $38.0$ & $-$ & $22.5$ & $27.9$ & $47.0$ \\
\midrule
VATT & $75.0$ & $36.5$ & & $63.5$ & $68.9$ & $\textbf{53.5}$ & $-$ & $22.3$ & $25.8$ & $49.9$ \\
VideoMAE &$73.6$ & $30.6$ & & $61.4$ & $76.0$ & $43.0$ & $-$ & $16.6$ & $23.3$ & $45.9$ \\
InternVideo & $75.5$ & $31.3$ & & $\textbf{63.9}$ & $73.6$ & $46.2$ & $-$ & $19.2$ & $25.5$ & $47.7$ \\
\bottomrule
\end{tabular}}
\end{center}
\end{table}

\textit{Generally speaking, DINOv2 performs the best in the frozen feature pooler evaluation (Tables~\ref{tab:frozen_table}), CLIP performs the best with multi-head attention pooler among image-native frozen FMs (Tables~\ref{tab:evl_table}), but CoCa catches up thanks to the low-rank adapter (Table~\ref{tab:adapter_table}).}
It is worth noting that this ranking of image-native frozen FMs differs from the ranking of image-native FMs in end-to-end finetuning. 
It seems that DINOv2 and CLIP's frozen features are more amendable to the video tasks than CoCa, but CoCa as a whole adapts better to video under both finetuning and the adapter. Hence, it is crucial to consider adaptation methods as an organic part of the evaluation of FMs to supply them various paths to demonstrate their capabilities.

\begin{figure}[t]
\centering
\begin{subfigure}{0.45\linewidth}
\centering
\includegraphics[height=0.5\linewidth]{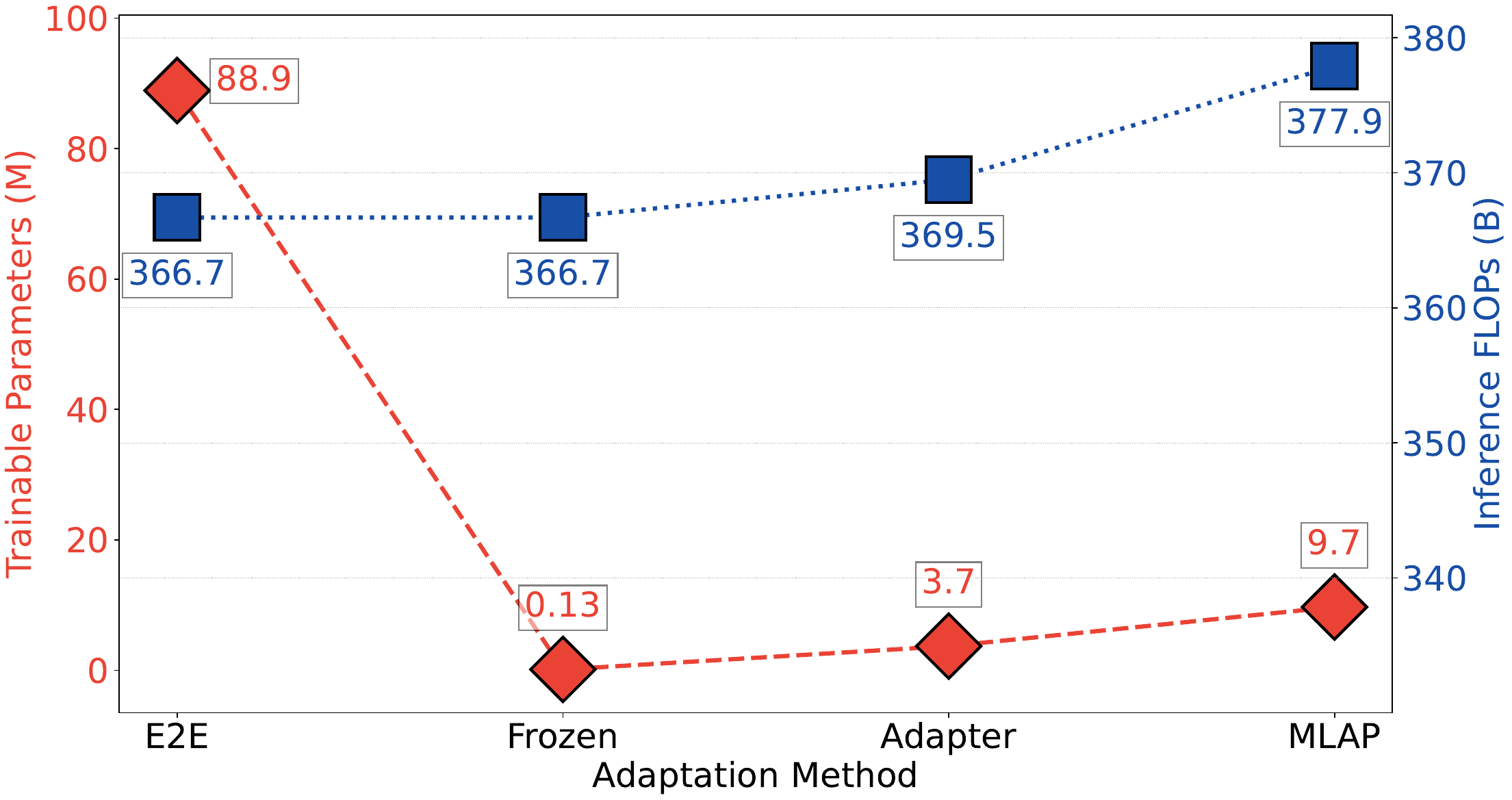}
\caption{}
\end{subfigure}
\quad
\begin{subfigure}{0.45\linewidth}
\centering
\includegraphics[height=0.5\linewidth]{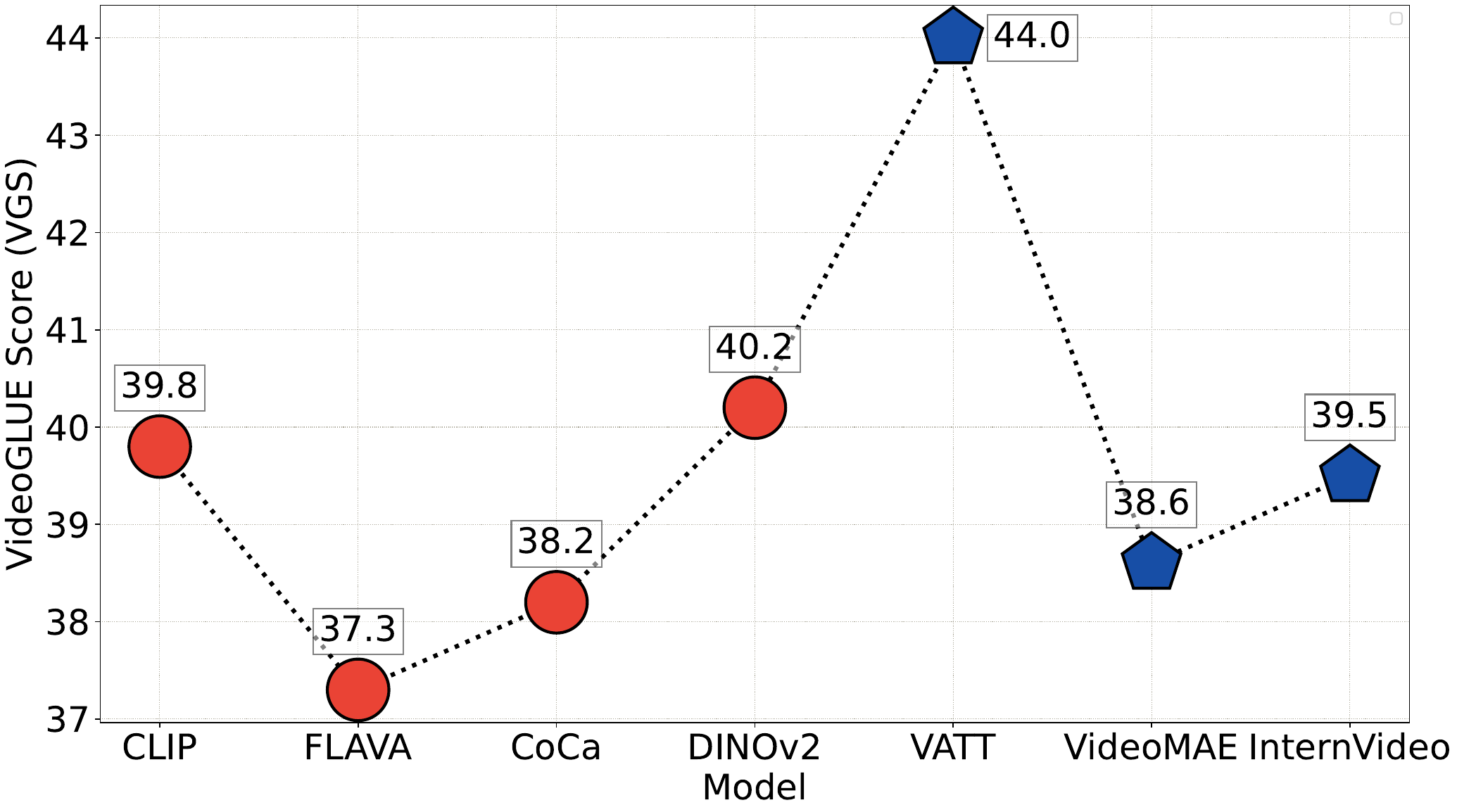}
\caption{}
\end{subfigure}
\vspace{-0.8em}
\caption{
(a) We measures the training (red diamond) and inference (blue square) cost of different adaptation methods in terms of number of trainable parameters and inference FLOPs, respectively. 
(b) We report \ourscore that combines a FM's performance weighted by its training costs with different adaptation methods for all the image-native (red circle) and video-native (blue pentagon) models.
}\label{fig:cost_and_vgs}
\end{figure}

\textit{Video-native FMs are better than image-native FMs in understanding motion-rich SSv2 and D48, Charades that contain multiple actions per video, and ANet for temporal action localization.} This observation is about the same as the one under end-to-end finetuning.
The image-native FMs are mainly superior on appearance-rich video datasets, where high-quality  spatial perceptual features are the key.
We conjecture that the vast image data empowering image-native FMs is more diverse in appearance than videos used to pretrain video-native FMs. 

\textit{Given frozen FMs, the low-rank adapter outperforms cross-attention layers, and multiple layers of cross-attention is better than a single cross-attention layer.}
Many works~\cite{caron2021dino,he2022masked} have shown features from different layers of a vision transformer have different attention maps. Hence, it is potentially beneficial to have an adaptation method to leverage multiple layers of a frozen FM. 
Table~\ref{tab:evl_table} reports the results with four cross-attention layers, whose average score per model (across different columns) is higher than that with a single cross-attention layer  (Table~\ref{tab:frozen_table}) by $18\%$ to $40\%$. The low-rank adapter (Table~\ref{tab:adapter_table}) further improves upon the cross-attention results partially because it explores all layers of a frozen FM.

\textit{On average, image-native FMs outperform video-native FMs under end-to-end  finetuning and the adapter, but it becomes the inverse in the other two adaptation methods.}
The adapter experiment paired with end-to-end finetuning experiment reveal the fact that {existing image-based FMs could be more easily adapted to video tasks when we could adjust the feature space of FMs}, possibly caused by the large-scale higher quality image(-text) pretraining datasets.
On the other hand, frozen feature experiments discussed above present us the inverse picture where video-based FMs perform better.
The seemingly paradox encourages more future research on bridging the gap on video-based pretraining with high-quality data, more effective modeling and better design on video benchmarks.

\subsection{Profiling foundation models for video understanding}

In this section, we consolidate our studies of the FMs with different adaptation methods and video tasks, focusing on their overall efficacy and efficiency. 
Specifically, we use trainable parameters and inference FLOPs to approximately represent the training and inference costs of an FM. 
Since all FMs in our evaluation are ViT-B and we align the same number of input tokens for each task.
The models have almost the same cost in one adaptation method. 
The left of Figure~\ref{fig:cost_and_vgs} shows the cost values for each adaptation method.
Note that an FM with LoRA adaptor tuning could have high inference cost despite lower training/adaptation costs than end-to-end fine-tuning.
While the figure provides a holistic view of an FM from multiple dimensions, one might be interested in a ranking among the FMs in terms of their video understanding capabilities. To this end, we summarize the multi-dimensional comparisons across different datasets, adaptation methods, and costs using a simplified scalar measure, termed \ourscore (\ourscoreabb), to probe an FM's general video understanding capability.

We use the cost values to normalize an adapted FM's average performance $s$ over all tasks.  
Formally, denoting by $\mathcal{S}_i$ an FM's average performance score over our video tasks under the $i$-th adaptation method and by $C_i^k$ the corresponding cost value under the $k$-th developmental scenario, we calculate the FM's \textit{\ourscoreabb}$^k$ by
\begin{equation}
    \textit{\ourscoreabb}^k = \sum_{i=1}^N w_i^k  \mathcal{S}_i, \; \text{where} \; w_i^k=\frac{\mathcal{A}_i^k}{\sum_{j=1}^N \mathcal{A}_j^k} \; \text{and} \; \mathcal{A}_i^k = \frac{1}{\log_{10}{C_i^k}},
\end{equation}
where $N=4$ is the number of adaptation methods, and $w_i\in[0,1]$ weighs score $\mathcal{S}_i$ according to the cost $C_i^k$. The final \ourscoreabb is the arithmetic average on $\{\text{\ourscoreabb}^k\}$, where $k=1,2$ corresponding to  training and inference respectively.

On the right panel of Figure~\ref{fig:cost_and_vgs}, we plot each FM's \ourscore.
We notice the average \ourscoreabb for video-native and image-native FMs on our video understanding tasks are $40.67$ vs.\ $38.84$ respectively.
Zooming in to the individual FMs, we find that VATT, a video-native FM, is at the first place with \ourscoreabb $43.97$, followed by the image-native DINOv2 with \ourscoreabb $40.17$.
This suggests that in-domain pretraining yields overall the best adaptation capability to video tasks, and  image-native FMs could also achieve competitive results on many but not all video understanding tasks. %

\section{Limitations}
\label{sec:limitations}
VideoGLUE serves as a comprehensive benchmark for studying and probing various video understanding capabilities of foundation models.
The current task portfolio includes various unimodal action understanding tasks.
We believe the scope of this work could be further extended as there are many other important video tasks not covered here, e.g. object or point-level tracking, long-term memory and forecasting.
Moreover, our benchmark could be strengthened by adding multimodal tasks like video captioning and question answering, given the rise of general Vision Language Models (VLM). 
We chose three representative FM adaptation methods and used them to provide as uniform experiment protocols for different FMs as possible. 
However, some of our observations could be flipped with the evolution of FMs development and adaptation methods, which are an active research area. 
We proposed a scalar score, \ourscore (\ourscoreabb), to capture the efficacy and efficiency of an FM on video understanding. 
However, \ourscoreabb might be dominated by one or a few datasets --- when it becomes a serious issue, we should probably improve the score and/or retire the datasets from future versions of VideoGLUE. 
Indeed, \ourscoreabb is not a perfect score that covers all aspects of FMs in a comprehensive manner. 
For example, it does not account for an FM's model size, model architecture, etc. 
We hope future research will lead to new metrics to complement \ourscoreabb and a more comprehensive evaluation of FMs for visual tasks.

\section{Conclusion}
\label{sec:conclusion}
In this report, we study four image-based and three video-based foundation models and their adaptation capability on general video understanding tasks.
Experiments are conducted on three hallmark video tasks, eight diverse datasets with four distinct adaption methods.
Our study shows existing image-based FMs performs well on some appearance-rich video datasets, while video-based FMs tend to achieve better on motion and temporal reasoning.
Four studied adaption methods curve different landscape, revealing the critical role of considering adaption methods as an organic part of evaluating FMs.
Finally, we propose one single metric \ourscoreabb to represent the video task adaptation efficiency of FMs.
We hope our research provides useful resources for evaluating and analyzing video foundation models, and address the current gap in foundation model evaluation within the video domain.

\section{Acknowledgement}
We would like to thank Xuhui Jia and Sergey Ioffe for reviewing and providing feedback on this paper.
We thank Albert Shaw on their early investigation on the UNIT architecture.
We also thank David Ross, Rahul Sukthankar, and Tomas Izo for their support and leadership on this project.

\appendix
\newpage
\setcounter{section}{0}
\renewcommand\thesection{\Alph{section}}
\renewcommand\thesubsection{\thesection.\arabic{subsection}}

\section*{\centering Supplementary Materials}
We first discuss the ethcial concerns and broader impact of this work (Section~\ref{app:ethical}).
We detail the datasets~(Section~\ref{app:datasets}), models~(Section~\ref{app:arch}), and training setups~(Section~\ref{app:hyperparameters}) in the supplementary materials to improve this work's reproducibility. Besides, Section~\ref{app:more-studies} includes more experimental studies to strengthen the main text.

\section{Ethical concern and broader impact}
\label{app:ethical}

\textbf{Ethical concern.}
We evaluate FMs on three video tasks, eight datasets in total. We select the tasks and datasets based on their popularity and representativeness. Although carefully designed, our benchmark inevitably inherited some ethical concerns from those datasets. For instance, many of the datasets are curated by crawling videos from the Internet, which do not proportionately represent the experiences of the global population and can potentially lead to biased evaluations of FMs. Moreover, the video datasets involve human daily activities, leading to  privacy concerns about the human actors in the videos. How to evaluate FMs for video understanding in a fair and privacy-preserving manner could be an important direction for future research.

\textbf{Broader impact.}
Our research reveals the need and tremendous opportunities to research video-first FMs by improving pretraining video data and methodologies. %
Our studies on different adaptation methods on versatile tasks confirms that both tasks and adaptation methods matter when it comes to the evaluation of FMs, shedding light on the already vibrant area of FM adaptations. %
Finally, we hope our research could inspire research on foundation models development and video understanding in general, along with their applications in the real world.

\section{Video understanding datasets} \label{app:datasets}

\begin{table}[h]
\caption{
Summary of dataset publishing year, venue and citations (as of October 17, 2024).
}
\label{tab:dataset_venue}
\begin{center}
\resizebox{0.8\linewidth}{!}{
\begin{tabular}{llccc} 
\toprule
Task & Dataset & Year & Venue & Citation \\
\midrule
\multirow{5}{*}{VC} & Kinetics-400~\cite{kay2017kinetics} & 2017 & arXiv & $4,558$ \\
& Moments in Time~\cite{mit} & 2018 & TPAMI & $651$ \\
& Something-Something v2~\cite{goyal2017something} & 2017 & ICCV & $1,582$ \\
& Diving48~\cite{li2018diving48} & 2018 & ECCV & $369$  \\
& Charades~\cite{sigurdsson2016charades} & 2016 & ECCV & $1,415$ \\
\midrule
TAL & ActivityNet v1.3~\cite{caba2015activitynet} & 2015 & CVPR & $2,881$ \\
\midrule
\multirow{2}{*}{STAL} & AVA v2.2~\cite{gu2018ava} & 2018 & CVPR & $1,200$ \\
& AVA-Kinetics~\cite{li2020avak} & 2020 & arXiv & $151$ \\
\bottomrule
\end{tabular}}
\end{center}
\end{table}

In Table~\ref{tab:dataset_venue} we show the publishing year, venues and citations to demonstrate the popularity and community acceptance of datasets of our choice. 
Below we provide dataset details.

\subsection{Appearance-focused action recognition}
Video classification is a task of classifying videos into pre-defined labels, with the major focus on human actions. 

Kinetics-400~\cite{kay2017kinetics} (K400) is a large-scale, high-quality video dataset widely used as a standard video classification benchmark.
It contains more than $250$K video clips with annotations of $400$ human daily actions. 
The actions are human focused and cover a broad range of classes including human-human interactions and human-object interactions.
Although the video clips span $10$ seconds on average, many studies~\cite{sevilla2021only, wang2018tsn} have pointed out the task could be easily solved on the Kinetics datasets by inferring from the static objects appeared or background environment --- motion information is less important than the visual appearance.
Hence, we categorize Kinetics400 as an appearance-focused action classification dataset.

Moments in Time~\cite{mit} (MiT) is a large-scale video event classification dataset, with one million human annotated short video clips (around $3$ seconds each). 
The temporal span corresponds to the averaged duration of human working memory and is a temporal envelope holding meaningful actions between people, objects, and phenomena.
Videos in MiT are annotated with 339 most used verbs in the English vocabulary. 

\subsection{Motion-focused action recognition}
Videos contain much more commonsense knowledge than still images do, such as an object's motion patterns and the causal consequences of an action, just to name a few.
However, appearance-based benchmarks do not evaluate a model's understanding of such commonsense knowledge, complex scenes, and situations. %
In observance of this, some video datasets have been proposed and studied in recent years with the focus on motions and common-sensing reasoning that are prosperous in video data. 

Something-Something v2~\cite{goyal2017something} (SSv2) is a collection of around $200$K videos of human performing pre-defined, basic actions with everyday objects. 
There are $174$ unique labels in total depicting atomic hand manipulations, like putting something into something, turning something upside down or covering something with something.
This dataset benchmarks a model's fine-grained understanding capability of object motions and scene changes by making the label space  atomic-action-focused and background-invariant.

Diving48~\cite{li2018diving48} (D48) is introduced to evaluate a model's dynamic reasoning capability. 
The video clips in this dataset are obtained by segmenting online videos of major diving competitions. In total, there are around $18$K videos annotated with $48$ classes. 
Because of its standardization, the diving scenario is purposefully chosen to avoid the scene, object, and person biases.

\subsection{Multi-label daily action classification}
Most of current action classification datasets involve video clips with a clean snapshot of a single action.
In contrast, humans perform daily complex activities step-by-step, simultaneously, or in an interleaving manner. 
Towards more comprehensive human daily activity reasoning, Charades~\cite{sigurdsson2016charades} is introduced. 
Different from web-collected datasets whose contents are more structured, Charades is collected by crowd-sourcing from hundreds of actors recording their videos in their own homes, acting out casual everyday activities.
Charades brings in more diversity into the video classification task due to its close-to-daily-life setting. 
Its videos are $30$ seconds long on average and have multi-label annotations testing models' understanding of complex daily activities with multiple steps.
Charades provides $110$k videos with $157$ action classes for training and evaluation.

\subsection{Temporal action localization}

Natural long videos contain scene changes and semantic shifts, while most of the existing video benchmarks formulate problems to focus on trimmed video clips.
Such a gap introduces evaluation bias as clip-level benchmarks could not reflect a model's temporal feature discriminativeness, which is of key importance to solve long-form video understanding tasks.
To comprehend the study on foundation models' video capabilities, we include the temporal action localization (TAL) task in our evaluation.
The task of TAL is to predict not only the action labels but also each action instance's temporal boundary in untrimmed videos.
We adopt ActivityNet v1.3~\cite{caba2015activitynet} as the dataset for the TAL task, which contains $10,002$  untrimmed videos in training and $4,985$ in validation.
The video length in this dataset is between $5$-$10$ minutes. In total, there are $200$ types of activities annotated.

\subsection{Spatiotemporal action localization}
Spatiotemporal Action Localization (STAL) is a person-centric task that asks a system to localize actors and predict their atomic actions~\cite{barker1955midwest, gu2018ava} in a transitory duration.

In AVA~\cite{gu2018ava}, $15$ minutes long movie clips are densely annotated at $1$Hz. In the key frames, every person is localized using a bounding box and labels corresponding to actions being performed by the actor.
The label vocabulary consists of $80$ different atomic visual actions. There are $430$ different movies in total.

AVA-Kinetics~\cite{li2020avak} follows the same labeling protocol as AVA, while its data source comes from the Kinetics700~\cite{kay2017kinetics} video pool.
The dataset contains over $230$K clips annotated with the $80$ AVA action classes for each of the humans in key frames.

\section{Model details} \label{app:arch}
\subsection{Task head architectures}
\begin{figure}[t]
\centering
\begin{subfigure}{0.195\linewidth}
\centering
\includegraphics[height=1.8\linewidth]{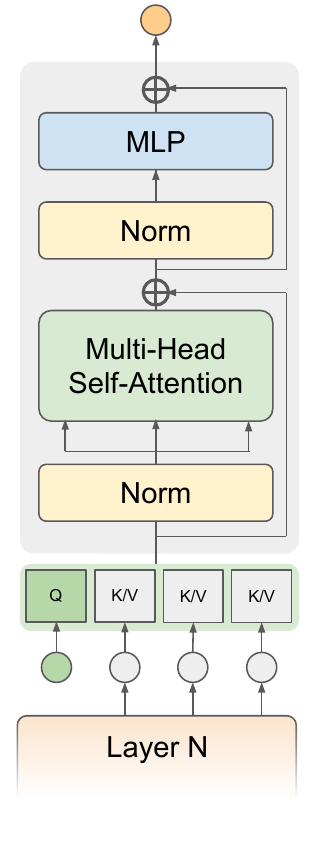}
\caption{}
\end{subfigure}
\begin{subfigure}{0.39\linewidth}
\centering
\includegraphics[height=0.9\linewidth]{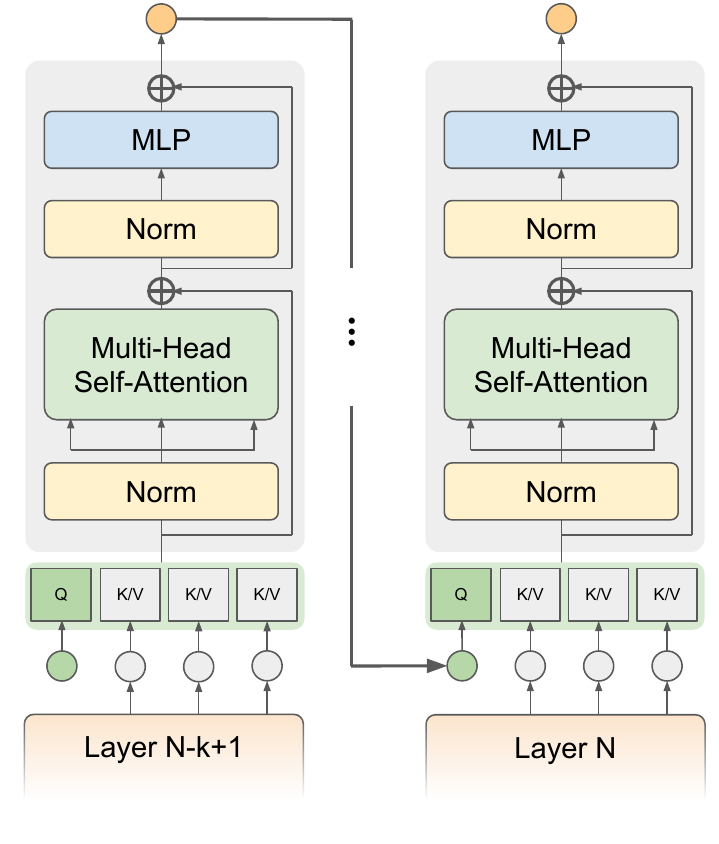}
\caption{}
\end{subfigure}
\begin{subfigure}{0.195\linewidth}
\centering
\includegraphics[height=1.8\linewidth]{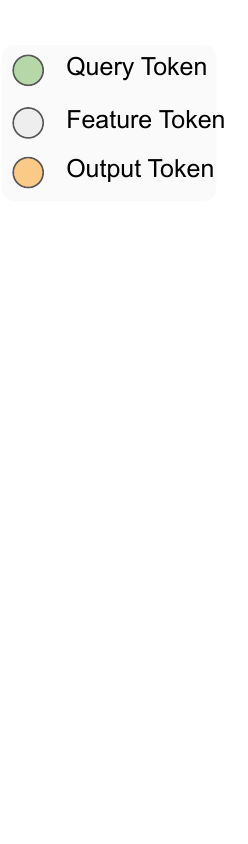}
\caption*{}
\end{subfigure}
\caption{
(a) Single-layer pooler head and (b) multi-layer attention pooling head for video classification and spatiotemporal action localization.
}\label{fig:arch_head}
\end{figure}

In Figure~\ref{fig:arch_head}, we plot the task heads used in our video classification and spatiotemporal action localization experiments, namely, the simple pooler head and multi-layer attention pooling head. 
For temporal localization, please refer to \citet{xu2019gtad} for the task head's detailed architecture.

\begin{figure}[t] 
    \centering
    \includegraphics[width=0.55\linewidth]{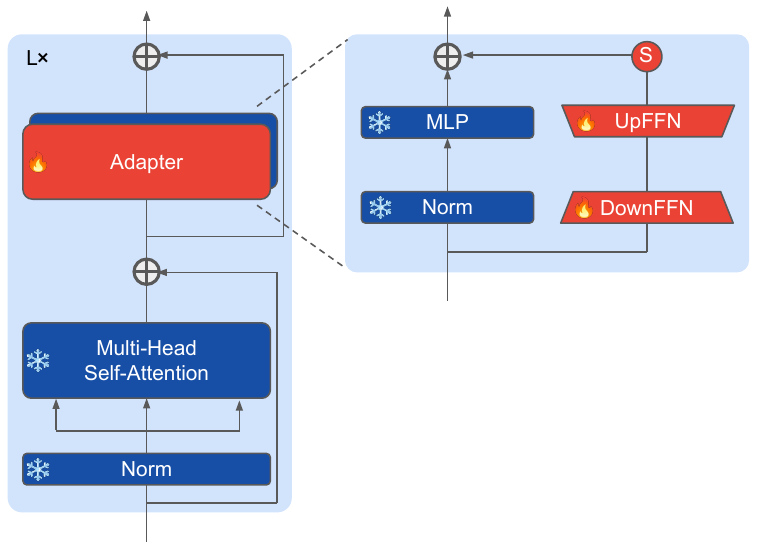}
    \vspace{-0.9em}
    \caption{The adapter used in vision transformer. 
    In the adapter layer, only the down-sample layer, up-sample layer, and the scaling factor are tunable.
    Between the down-sample layer and up-sample layer, an activation function is applied, which in our case is ReLU.
    }
    \label{fig:arch_adapter}
\end{figure}

Figure~\ref{fig:arch_adapter} illustrates the encoder adapter layer's architecture. In the the adapter layer, only the down-sample layer, up-sample layer, and the scaling factor are tunable.

\subsection{Image-to-video adaptation}

\begin{table}[!t]
\caption{Early vs.\ late fusion on image-native FMs. In this experiment, the frozen feature with a single-layer pooler head is used.}
\label{tab:i2v_early_vs_late}
\begin{center}
\begin{tabular}{lccccc}
\toprule
    
\multirow{2}{*}{Models}  & \multicolumn{2}{c}{K400} & & \multicolumn{2}{c}{SSv2}   \\
\cline{2-3} \cline{5-6}
&  Early & Late & & Early & Late \\

\midrule
CLIP & $70.5$& $75.2$ & &$38.1$ &$41.0$\\
FLAVA & $67.9$& $71.3$ & &$40.4$ &$40.6$\\
CoCa & $72.7$& $61.4$ & &$41.5$ &$33.3$\\
\bottomrule
\end{tabular}
\end{center}
\end{table}

\begin{table}[!t]
\caption{
Ablation study on the temporal positional embedding for image-to-video adaption.
We choose FLAVA~\cite{singh2022flava} with the frozen feature setting in this experiment.}
\label{tab:i2v_flava_temporal}
\begin{center}
\begin{tabular}{ccccccc}
\toprule
 
Adds temporal& \multicolumn{2}{c}{VC-A} & & \multicolumn{2}{c}{VC-M} & VC-ML \\
\cline{2-3} \cline{5-6}
positional embedding? & K400 & MiT & & SSv2 & D48 & Charades
\\
\midrule
\xmark & $71.3$ & $29.7$ & & $30.3$ & $41.6$ & $10.7$ 
\\
\cmark & $71.3$ & $29.7$ & & $40.6$ & $45.9$ & $12.6$ 
\\
\bottomrule
\end{tabular}
\end{center}
\end{table}

Adapting image backbones to video tasks requires us to fuse the image embeddings at some point in the network and also introduce additional temporal information. 

We consider two choices, early-fusion and late-fusion, and ablate them in the frozen feature setting in Table~\ref{tab:i2v_early_vs_late}. 
In both early-fusion and late-fusion, we first apply the projection layer on each frame independently to embed pixel patches into embedding tokens.
We then average-pool the embedding tokens from nearby frames to reduce the sequence length to $n \times h \times w$. %
In the early-fusion setting, we pass all tokens \emph{together} to the image backbone to extract video features.
In late-fusion, we pass each set of $h \times w$ tokens \emph{independently} to the image backbone.
Empirically, we find that the FLAVA~\cite{singh2022flava} and CLIP~\cite{radford2021clip} models do better with late-fusion while CoCa~\cite{yu2022coca} does better with early-fusion. 

Furthermore, we ablate the importance of temporal information using the frozen-features from FLAVA~\cite{singh2022flava}.
In Table~\ref{tab:i2v_flava_temporal}, we find that adding temporal positional embedding to the input is essential for D48~\cite{li2018diving48},  SSv2~\cite{goyal2017something}, and Charades~\cite{sigurdsson2016charades} while not necessary for K400~\cite{kay2017kinetics} and MiT~\cite{mit}. 
This supports our grouping that K400 and MiT are appearance-focused datasets.

Based on these findings, we use late-fusion for FLAVA~\cite{singh2022flava} and CLIP~\cite{radford2021clip}  and early-fusion for CoCa~\cite{yu2022coca}. %
We add learnable temporal positional embeddings for all the image-native FMs.

\section{Task-specific hyperparameters} \label{app:hyperparameters}
In the following, we provide experiment settings and hyperparamters we used in this study.
In Table~\ref{tab:vc_configs}, we list the hyperparameters we applied in the video classification task.
In Table~\ref{tab:stal_configs}, we present the hyperparameters we used on spatiotemporal action localization.
In Table~\ref{tab:tal_configs}, we present the hyperparameters we used on temporal action localization task.

We performed a greedy search on the learning rate and weight decay in all our experiments while keeping most other hyperparameters (e.g., data augmentation magnitude, dropout rate, drop path rate, etc.) consistent across different models and datasets. Specifically, we start with learning rate $1$e-$4$ and weight decay $1$e-$5$ and uniformly sample learning rates and weight decay factors with a rate of $5$ and $10$, respectively, centered around the starting points. After the first round, we pick the best-identified learning rate and weight decay factor as the new starting point and conduct another round of sampling with a rate of $2$. 
We repeat another two to three rounds of hyperparameter search (with a rate of $2$) until the model's performance converges. This process is a trade-off between computation costs and thoroughly examining an FM's performance under each experiment setup. 
The search ranges for the learning rate and weight decay are [$4$e-$5$, $2.5$e-$3$] and [$1$e-$6$, $1$e-$4$], respectively. 
We found that the learning rate is the most crucial factor when adapting an FM to downstream video understanding tasks. 

\begin{table}[!t]
    \caption{Experimental configurations for video classification tasks. We let learning rate and weight decay to be tunable per model to allow some flexibility for task adaptations.}
	\label{tab:vc_configs}
	\begin{center}
    \resizebox{\linewidth}{!}{
        \begin{tabular}	{llllll}
			\toprule
            Configurations & K400 & MiT & SSv2 & D48 & Charades \\
			\midrule
    \textbf{General}\\
	Batch size & 256  & 256 & 256 & 256 & 256 \\
	Training epochs & 150 & 50 & 50 & 100 & 50 \\
	ViT sequence length & 8 $\times$ 14 $\times$ 14  & 8 $\times$ 14 $\times$ 14 & 8 $\times$ 14 $\times$ 14 & 8 $\times$ 14 $\times$ 14 & 8 $\times$ 14 $\times$ 14 \\
	\hline
    \textbf{Optimization}\\
	Optimizer & AdamW  & AdamW & AdamW & AdamW & AdamW \\
	Optimizer momentum & 0.9 & 0.9 & 0.9 & 0.9 & 0.9\\
	Learning rate schedule & Cosine decay  & Cosine decay & Cosine decay  & Cosine decay  & Cosine decay \\
	Warmup ratio & 5\%  & 5\% & 5\% & 5\% & 5\% \\
	\hline
    \textbf{Data augmentations}\\
	Random horizontal flip & Yes & Yes & No & Yes & No \\
	Aspect ratio & (0.5, 2.0) & (0.5, 2.0)  & (0.5, 2.0) & (0.5, 2.0) &  (0.5, 2.0) \\
	Area ratio & (0.3, 1.0)  & (0.3, 1.0)  & (0.3, 1.0)  & (0.3, 1.0) &  (0.3, 1.0) \\
	RandAug & (9, 0.5) & - & (9, 0.5) & - & - \\
	MixUp & 0.8 & - & 0.8 & - & -\\
	CutMix & 1.0 & - & 1.0 & - & -\\
	\hline
    \textbf{Evaluation}\\
    Multi-clips & 4 & 4 & 1 & 4 & 4 \\
    Multi-views & 3 & 3 & 3 & 3 & 3 \\
    Segment-based sampling & No & No & Yes & No & No \\
			\bottomrule
		\end{tabular}}
	\end{center}
\end{table}

\begin{table}[!t]
    \caption{Experimental configurations for spatiotemporal action localization.}
	\label{tab:stal_configs}
	\begin{center}
    \resizebox{0.6\linewidth}{!}{
		\begin{tabular}	{lll}
			\toprule
			Configurations & AVA v2.2 & AVA-Kinetics \\
			\midrule
	\textbf{General} \\
	Batch size & 256  & 256  \\
	Training epochs & 50 & 50  \\
	ViT sequence length & 8 $\times$ 16 $\times$ 16  & 8 $\times$ 16 $\times$ 16 \\
	\hline
	\textbf{Optimization} \\
	Optimizer & AdamW  & AdamW \\ 
	Optimizer momentum & 0.9 & 0.9 \\
	Layer decay & 0.75  & 0.75 \\ 
	Learning rate schedule & Cosine decay  & Cosine decay \\
	Warmup ratio & 5\%  & 5\% \\
	\hline
	\textbf{Data augmentations} \\
	Random horizontal flip & Yes & Yes \\
	Random scale & (0.5, 2.0) & (0.5, 2.0) \\
	Random color augmentation & Yes & Yes \\
		\bottomrule
		\end{tabular}}
	\end{center}
\end{table}

\begin{table}[!t]
    \caption{Experimental configurations for temporal action localization.}
	\label{tab:tal_configs}
	\begin{center}
    \resizebox{0.45\linewidth}{!}{
		\begin{tabular}	{ll}
			\toprule
			Configurations & ActivityNet v1.3 \\
			\midrule
	\textbf{General} \\
	Batch size & 32 \\
	Training epochs & 10 \\
	\hline
	\textbf{Feature extraction} \\
    Frame rate (FPS) & 15 \\
	Per-clip length (second) & 16 \\
	Clip stride & 16 \\
	\hline
	\textbf{Optimization} \\
	Optimizer & AdamW \\ 
	Optimizer momentum & 0.9 \\
	Learning rate schedule & Cosine decay \\
	\bottomrule
	\end{tabular}}
	\end{center}
\end{table}

\section{More studies}\label{app:more-studies}
\subsection{Large model adaptations}

\begin{table}[!t]
\caption{
Evaluating large-scale FMs when using frozen feature with a one-layer pooler head.
We report the Top-1 accuracy on K400, MiT, D48, SSv2 and MAP on Charades.
}
\label{tab:large_model_table}
\begin{center}
\resizebox{0.65\linewidth}{!}{
\begin{tabular}{lccccccc}
\toprule
 
 \multirow{2}{*}{Models}& & \multicolumn{2}{c}{VC-A} & & \multicolumn{2}{c}{VC-M} & VC-ML \\
\cline{3-4} \cline{6-7}
& & K400 & MiT & & SSv2 & D48 & Charades \\
\toprule
InternVideo-L &  & $78.6$ & $33.7$ & & $67.4$ & $69.6$  & $20.9$  \\
VideoMAE-v2-B/DL &  & $86.7$ & $38.9$ & & $57.7$ & $61.4$ & $33.2$ \\
VideoMAE-v2-g & & $59.7$ & $20.7$ & & $44.2$ & $42.5$ & $12.7$ \\
VideoMAE-v2-g/FT & & $82.1$ & $35.0$ & & $56.1$ & $60.5$ & $22.4$ \\
InternVideo-v2-g & & $85.0$ & $43.0$ & & $61.6$ & $53.1$ & $40.9$ \\
VideoPrism-g & & $86.6$ & $44.7$ & & $67.4$ & $66.1$ & $61.0$ \\
\bottomrule
\end{tabular}}
\end{center}
\end{table}

For the completeness of this report and reader's reference, in Table~\ref{tab:large_model_table} we report experimental results under our settings with large FMs under the frozen backbone with one pooler head setup.

VideoMAE-v2-B/DL~\cite{wang2023videomaev2} denotes the ViT-B model distilled from ViT-g on the Kinetics710 datasets\footnote{\url{https://github.com/OpenGVLab/VideoMAEv2/blob/master/docs/MODEL_ZOO.md}}. 
VideoMAE-v2-g~\cite{wang2023videomaev2} is the model that pretrained on UnlabeledHybrid dataset, while VideoMAE-v2-g/FT~\cite{wang2023videomaev2} conducts further finetuning using supervised training on Kinetics710.
InternVideo-v2-g~\cite{wang2024internvideo2} and VideoPrism-g~\cite{zhao2024videoprism} are two video foundation models with multi-stage pre-training on curated in-house web video data.
For InternVideo-v2-g, we use their stage-2 checkpoint\footnote{\url{https://github.com/OpenGVLab/InternVideo/blob/main/InternVideo2/multi_modality/MODEL_ZOO.md}}.
For videoPrism-g, we use their final checkpoint.

\subsection{Sample-efficient transfer learning}

\begin{table}[t!]
\caption{
Benchmark FMs adaptation on video understanding tasks under sample-efficient transfer learning.
This table shows Top-1 classification accuracy and the relative accuracy to training with $100\%$ data (shown in parentheses). 
Results are achieved by using frozen features with pooler head.
}
\label{tab:few_shot}
\begin{center}
\resizebox{0.88\linewidth}{!}{
\begin{tabular}{lccccccc}
\toprule
    
 \multirow{2}{*}{Models} & \multicolumn{3}{c}{K400} & &\multicolumn{3}{c}{SSv2} \\
\cline{2-4} \cline{6-8}
& $1\%$ &  $10\%$ & $100\%$ & & $1\%$ & $10\%$ & $100\%$ \\

\midrule
CLIP & $36.9$ \footnotesize{$(46.2\%)$} &$66.8$ \footnotesize{$(83.6\%)$} &$79.0$ & &$8.7$ \footnotesize{$(19.3\%)$} &$25.1$ \footnotesize{$(55.5\%)$} &$45.3$ \\
FLAVA &$14.4$ \footnotesize{$(20.2\%)$} &$35.8$ \footnotesize{$ (50.3\%)$} &$71.3$ & &$7.2$ \footnotesize{$(17.7\%)$} &$14.3$ \footnotesize{$(35.3\%)$} &$40.6$ \\
CoCa & $27.1$ \footnotesize{$(37.8\%)$}& $48.9$ \footnotesize{$(67.0\%)$} &$73.1$ & &$5.6$ \footnotesize{$(13.4\%)$} &$20.9$ \footnotesize{$(50.4\%)$} &$41.5$ \\
\midrule
VATT &$34.1$ \footnotesize{$(45.4\%)$} &$63.7$ \footnotesize{$(84.8\%)$} &$75.1$ & &$12.9$ \footnotesize{$(22.4\%)$} &$37.6$ \footnotesize{$(65.0\%)$} &$57.8$\\
VideoMAE &$15.5$ \footnotesize{$(23.9\%)$} &$32.0$ \footnotesize{$(49.2\%)$} &$65.0$ & & $13.7$ \footnotesize{$(25.4\%)$} & $30.3$ \footnotesize{$(56.2\%)$} & $53.9$ \\
InternVideo & $20.4$ \footnotesize{$(29.5\%)$} & $50.2$ \footnotesize{$(72.4\%)$} &$69.3$ & & $19.5$ \footnotesize{$(33.6\%)$} & $41.1$ \footnotesize{$(70.7\%)$} &$58.2$\\
\bottomrule
\end{tabular}}
\end{center}
\end{table}

A strong FM should be able to adapt to downstream tasks with a few training samples.
In this section, we test the adaption ability of FMs in a sample-efficient transfer learning setting.
Particularly, we freeze backbones and train a pooler head to adapt the FMs on K400 and SSv2.
For either dataset, we sample $1\%$ and $10\%$ data from the training set uniformly for training and evaluate on the full evaluation dataset.

We show our experimental results in Table~\ref{tab:few_shot}.
To better understand the data efficiency, we also show the relative Top-1 accuracy for each model (shown in the bracket), which is defined as the ratio between accuracy with fewer training examples and the accuracy achieved using all the training data. 
A higher relative Top-1 accuracy means the performance of the model is closer to its ``full'' capacity under the sample-efficient setting.
We notice that the best performed model on each dataset in fully fine-tuned model also performs best in the few-shot setting.
Especially, CLIP~\cite{radford2021clip} achieves $46.2\%$ and $83.6\%$ relative Top-1 accuracy on K400 using only $1\%$ and $10\%$ of the training data, respectively.
On SSv2, InternVideo~\cite{wang2022internvideo} achieves $33.6\%$ and $70.6\%$ relative Top-1 accuracy with only $1\%$ and $10\%$ of the training data.

\clearpage
{\small
\bibliographystyle{tmlr}
\bibliography{my_bib}

\begin{thebibliography}{67}
\providecommand{\natexlab}[1]{#1}
\providecommand{\url}[1]{\texttt{#1}}
\expandafter\ifx\csname urlstyle\endcsname\relax
  \providecommand{\doi}[1]{doi: #1}\else
  \providecommand{\doi}{doi: \begingroup \urlstyle{rm}\Url}\fi

\bibitem[Akbari et~al.(2021)Akbari, Yuan, Qian, Chuang, Chang, Cui, and
  Gong]{akbari2021vatt}
Hassan Akbari, Liangzhe Yuan, Rui Qian, Wei-Hong Chuang, Shih-Fu Chang, Yin
  Cui, and Boqing Gong.
\newblock {VATT: Transformers for multimodal self-supervised learning from raw
  video, audio and text}.
\newblock In \emph{NeurIPS}, 2021.

\bibitem[Alayrac et~al.(2022)Alayrac, Donahue, Luc, Miech, Barr, Hasson, Lenc,
  Mensch, Millican, Reynolds, et~al.]{alayrac2022flamingo}
Jean-Baptiste Alayrac, Jeff Donahue, Pauline Luc, Antoine Miech, Iain Barr,
  Yana Hasson, Karel Lenc, Arthur Mensch, Katherine Millican, Malcolm Reynolds,
  et~al.
\newblock Flamingo: a visual language model for few-shot learning.
\newblock In \emph{NeurIPS}, 2022.

\bibitem[Alwassel et~al.(2021)Alwassel, Giancola, and Ghanem]{alwassel2021tsp}
Humam Alwassel, Silvio Giancola, and Bernard Ghanem.
\newblock {TSP: Temporally-sensitive pretraining of video encoders for
  localization tasks}.
\newblock In \emph{ICCV}, 2021.

\bibitem[Anil et~al.(2023)Anil, Dai, Firat, Johnson, Lepikhin, Passos, Shakeri,
  Taropa, Bailey, Chen, et~al.]{anil2023palm}
Rohan Anil, Andrew~M Dai, Orhan Firat, Melvin Johnson, Dmitry Lepikhin,
  Alexandre Passos, Siamak Shakeri, Emanuel Taropa, Paige Bailey, Zhifeng Chen,
  et~al.
\newblock {PaLM 2 technical report}.
\newblock \emph{arXiv preprint arXiv:2305.10403}, 2023.

\bibitem[Bao et~al.(2021)Bao, Dong, Piao, and Wei]{bao2021beit}
Hangbo Bao, Li~Dong, Songhao Piao, and Furu Wei.
\newblock {BEIT: BERT pre-training of image transformers}.
\newblock \emph{arXiv preprint arXiv:2106.08254}, 2021.

\bibitem[Barker \& Wright(1955)Barker and Wright]{barker1955midwest}
Roger~G Barker and Herbert~F Wright.
\newblock {Midwest and its children: The psychological ecology of an American
  town}.
\newblock \emph{Marriage and Family Living}, 1955.

\bibitem[Bommasani et~al.(2021)Bommasani, Hudson, Adeli, Altman, Arora, von
  Arx, Bernstein, Bohg, Bosselut, Brunskill, et~al.]{foundation-model}
Rishi Bommasani, Drew~A Hudson, Ehsan Adeli, Russ Altman, Simran Arora, Sydney
  von Arx, Michael~S Bernstein, Jeannette Bohg, Antoine Bosselut, Emma
  Brunskill, et~al.
\newblock {On the opportunities and risks of foundation models}.
\newblock \emph{arXiv preprint arXiv:2108.07258}, 2021.

\bibitem[Brown et~al.(2020)Brown, Mann, Ryder, Subbiah, Kaplan, Dhariwal,
  Neelakantan, Shyam, Sastry, Askell, et~al.]{brown2020gpt3}
Tom Brown, Benjamin Mann, Nick Ryder, Melanie Subbiah, Jared~D Kaplan, Prafulla
  Dhariwal, Arvind Neelakantan, Pranav Shyam, Girish Sastry, Amanda Askell,
  et~al.
\newblock {Language models are few-shot learners}.
\newblock In \emph{NeurIPS}, 2020.

\bibitem[Buch et~al.(2022)Buch, Eyzaguirre, Gaidon, Wu, Fei-Fei, and
  Niebles]{buch2022cvpr}
Shyamal Buch, Cristobal Eyzaguirre, Adrien Gaidon, Jiajun Wu, Li~Fei-Fei, and
  Juan~Carlos Niebles.
\newblock {Revisiting the ``video'' in video-language understanding}.
\newblock In \emph{CVPR}, 2022.

\bibitem[Caron et~al.(2021)Caron, Touvron, Misra, J{\'e}gou, Mairal,
  Bojanowski, and Joulin]{caron2021dino}
Mathilde Caron, Hugo Touvron, Ishan Misra, Herv{\'e} J{\'e}gou, Julien Mairal,
  Piotr Bojanowski, and Armand Joulin.
\newblock {Emerging properties in self-supervised vision transformers}.
\newblock In \emph{ICCV}, 2021.

\bibitem[Chen et~al.(2022)Chen, Wang, Changpinyo, Piergiovanni, Padlewski,
  Salz, Goodman, Grycner, Mustafa, Beyer, et~al.]{chen2022pali}
Xi~Chen, Xiao Wang, Soravit Changpinyo, AJ~Piergiovanni, Piotr Padlewski,
  Daniel Salz, Sebastian Goodman, Adam Grycner, Basil Mustafa, Lucas Beyer,
  et~al.
\newblock {PaLI: A jointly-scaled multilingual language-image model}.
\newblock \emph{arXiv preprint arXiv:2209.06794}, 2022.

\bibitem[Chowdhery et~al.(2022)Chowdhery, Narang, Devlin, Bosma, Mishra,
  Roberts, Barham, Chung, Sutton, Gehrmann, et~al.]{chowdhery2022palm}
Aakanksha Chowdhery, Sharan Narang, Jacob Devlin, Maarten Bosma, Gaurav Mishra,
  Adam Roberts, Paul Barham, Hyung~Won Chung, Charles Sutton, Sebastian
  Gehrmann, et~al.
\newblock {PaLM: Scaling language modeling with pathways}.
\newblock \emph{arXiv preprint arXiv:2204.02311}, 2022.

\bibitem[Devlin et~al.(2018)Devlin, Chang, Lee, and Toutanova]{devlin2018bert}
Jacob Devlin, Ming-Wei Chang, Kenton Lee, and Kristina Toutanova.
\newblock {BERT: Pre-training of deep bidirectional transformers for language
  understanding}.
\newblock \emph{arXiv preprint arXiv:1810.04805}, 2018.

\bibitem[Dong et~al.(2019)Dong, Yang, Wang, Wei, Liu, Wang, Gao, Zhou, and
  Hon]{dong2019unified}
Li~Dong, Nan Yang, Wenhui Wang, Furu Wei, Xiaodong Liu, Yu~Wang, Jianfeng Gao,
  Ming Zhou, and Hsiao-Wuen Hon.
\newblock {Unified language model pre-training for natural language
  understanding and generation}.
\newblock In \emph{NeurIPS}, 2019.

\bibitem[Dosovitskiy et~al.(2020)Dosovitskiy, Beyer, Kolesnikov, Weissenborn,
  Zhai, Unterthiner, Dehghani, Minderer, Heigold, Gelly,
  et~al.]{dosovitskiy2020vit}
Alexey Dosovitskiy, Lucas Beyer, Alexander Kolesnikov, Dirk Weissenborn,
  Xiaohua Zhai, Thomas Unterthiner, Mostafa Dehghani, Matthias Minderer, Georg
  Heigold, Sylvain Gelly, et~al.
\newblock {An image is worth 16x16 words: Transformers for image recognition at
  scale}.
\newblock \emph{arXiv preprint arXiv:2010.11929}, 2020.

\bibitem[Fabian Caba~Heilbron \& Niebles(2015)Fabian Caba~Heilbron and
  Niebles]{caba2015activitynet}
Bernard~Ghanem Fabian Caba~Heilbron, Victor~Escorcia and Juan~Carlos Niebles.
\newblock {ActivityNet: A large-scale video benchmark for human activity
  understanding}.
\newblock In \emph{CVPR}, 2015.

\bibitem[Fan et~al.(2021)Fan, Xiong, Mangalam, Li, Yan, Malik, and
  Feichtenhofer]{fan2021multiscale}
Haoqi Fan, Bo~Xiong, Karttikeya Mangalam, Yanghao Li, Zhicheng Yan, Jitendra
  Malik, and Christoph Feichtenhofer.
\newblock {Multiscale vision transformers}.
\newblock In \emph{ICCV}, 2021.

\bibitem[Feichtenhofer et~al.(2019)Feichtenhofer, Fan, Malik, and
  He]{feichtenhofer2019slowfast}
Christoph Feichtenhofer, Haoqi Fan, Jitendra Malik, and Kaiming He.
\newblock {SlowFast networks for video recognition}.
\newblock In \emph{ICCV}, 2019.

\bibitem[Feichtenhofer et~al.(2021)Feichtenhofer, Fan, Xiong, Girshick, and
  He]{feichtenhofer2021large}
Christoph Feichtenhofer, Haoqi Fan, Bo~Xiong, Ross Girshick, and Kaiming He.
\newblock {A large-scale study on unsupervised spatiotemporal representation
  learning}.
\newblock In \emph{CVPR}, 2021.

\bibitem[Feichtenhofer et~al.(2022)Feichtenhofer, Fan, Li, and
  He]{feichtenhofer2022vmae}
Christoph Feichtenhofer, Haoqi Fan, Yanghao Li, and Kaiming He.
\newblock {Masked autoencoders as spatiotemporal learners}.
\newblock \emph{arXiv preprint arXiv:2205.09113}, 2022.

\bibitem[Goyal et~al.(2017)Goyal, Ebrahimi~Kahou, Michalski, Materzynska,
  Westphal, Kim, Haenel, Fruend, Yianilos, Mueller-Freitag,
  et~al.]{goyal2017something}
Raghav Goyal, Samira Ebrahimi~Kahou, Vincent Michalski, Joanna Materzynska,
  Susanne Westphal, Heuna Kim, Valentin Haenel, Ingo Fruend, Peter Yianilos,
  Moritz Mueller-Freitag, et~al.
\newblock {The ``something something'' video database for learning and
  evaluating visual common sense}.
\newblock In \emph{ICCV}, 2017.

\bibitem[Gu et~al.(2018)Gu, Sun, Ross, Vondrick, Pantofaru, Li,
  Vijayanarasimhan, Toderici, Ricco, Sukthankar, et~al.]{gu2018ava}
Chunhui Gu, Chen Sun, David~A Ross, Carl Vondrick, Caroline Pantofaru, Yeqing
  Li, Sudheendra Vijayanarasimhan, George Toderici, Susanna Ricco, Rahul
  Sukthankar, et~al.
\newblock {AVA: A video dataset of spatio-temporally localized atomic visual
  actions}.
\newblock In \emph{CVPR}, 2018.

\bibitem[He et~al.(2022)He, Chen, Xie, Li, Doll{\'a}r, and
  Girshick]{he2022masked}
Kaiming He, Xinlei Chen, Saining Xie, Yanghao Li, Piotr Doll{\'a}r, and Ross
  Girshick.
\newblock {Masked autoencoders are scalable vision learners}.
\newblock In \emph{CVPR}, 2022.

\bibitem[Hu et~al.(2021)Hu, Shen, Wallis, Allen-Zhu, Li, Wang, Wang, and
  Chen]{hu2021lora}
Edward~J Hu, Yelong Shen, Phillip Wallis, Zeyuan Allen-Zhu, Yuanzhi Li, Shean
  Wang, Lu~Wang, and Weizhu Chen.
\newblock {LoRA: Low-rank adaptation of large language models}.
\newblock In \emph{ICLR}, 2021.

\bibitem[Huang et~al.(2023)Huang, Dong, Wang, Hao, Singhal, Ma, Lv, Cui,
  Mohammed, Liu, et~al.]{huang2023language}
Shaohan Huang, Li~Dong, Wenhui Wang, Yaru Hao, Saksham Singhal, Shuming Ma,
  Tengchao Lv, Lei Cui, Owais~Khan Mohammed, Qiang Liu, et~al.
\newblock {Language is not all you need: Aligning perception with language
  models}.
\newblock \emph{arXiv preprint arXiv:2302.14045}, 2023.

\bibitem[Jaderberg et~al.(2015)Jaderberg, Simonyan, Zisserman,
  et~al.]{jaderberg2015spatial_transformer}
Max Jaderberg, Karen Simonyan, Andrew Zisserman, et~al.
\newblock {Spatial transformer networks}.
\newblock In \emph{NeurIPS}, 2015.

\bibitem[Jia et~al.(2021)Jia, Yang, Xia, Chen, Parekh, Pham, Le, Sung, Li, and
  Duerig]{jia2021align}
Chao Jia, Yinfei Yang, Ye~Xia, Yi-Ting Chen, Zarana Parekh, Hieu Pham, Quoc Le,
  Yun-Hsuan Sung, Zhen Li, and Tom Duerig.
\newblock {Scaling up visual and vision-language representation learning with
  noisy text supervision}.
\newblock In \emph{ICML}, 2021.

\bibitem[Ju et~al.(2022)Ju, Han, Zheng, Zhang, and Xie]{ju2022prompting}
Chen Ju, Tengda Han, Kunhao Zheng, Ya~Zhang, and Weidi Xie.
\newblock {Prompting visual-language models for efficient video understanding}.
\newblock In \emph{ECCV}, 2022.

\bibitem[Kay et~al.(2017)Kay, Carreira, Simonyan, Zhang, Hillier,
  Vijayanarasimhan, Viola, Green, Back, Natsev, et~al.]{kay2017kinetics}
Will Kay, Joao Carreira, Karen Simonyan, Brian Zhang, Chloe Hillier, Sudheendra
  Vijayanarasimhan, Fabio Viola, Tim Green, Trevor Back, Paul Natsev, et~al.
\newblock {The Kinetics human action video dataset}.
\newblock \emph{arXiv preprint arXiv:1705.06950}, 2017.

\bibitem[Kondratyuk et~al.(2021)Kondratyuk, Yuan, Li, Zhang, Tan, Brown, and
  Gong]{kondratyuk2021movinets}
Dan Kondratyuk, Liangzhe Yuan, Yandong Li, Li~Zhang, Mingxing Tan, Matthew
  Brown, and Boqing Gong.
\newblock {MoviNets: Mobile video networks for efficient video recognition}.
\newblock In \emph{CVPR}, 2021.

\bibitem[Lei et~al.(2023)Lei, Berg, and Bansal]{jie23acl}
Jie Lei, Tamara~L. Berg, and Mohit Bansal.
\newblock {Revealing single frame bias for video-and-language learning}.
\newblock In \emph{ACL}, 2023.

\bibitem[Li et~al.(2020)Li, Thotakuri, Ross, Carreira, Vostrikov, and
  Zisserman]{li2020avak}
Ang Li, Meghana Thotakuri, David~A Ross, Jo{\~a}o Carreira, Alexander
  Vostrikov, and Andrew Zisserman.
\newblock {The AVA-Kinetics localized human actions video dataset}.
\newblock \emph{arXiv preprint arXiv:2005.00214}, 2020.

\bibitem[Li et~al.(2022{\natexlab{a}})Li, Liu, Li, Zhang, Aneja, Yang, Jin,
  Lee, Hu, Liu, et~al.]{benchmark-microsoft}
Chunyuan Li, Haotian Liu, Liunian~Harold Li, Pengchuan Zhang, Jyoti Aneja,
  Jianwei Yang, Ping Jin, Yong~Jae Lee, Houdong Hu, Zicheng Liu, et~al.
\newblock {ELEVATER: A benchmark and toolkit for evaluating Language-Augmented
  Visual Models}.
\newblock \emph{arXiv preprint arXiv:2204.08790}, 2022{\natexlab{a}}.

\bibitem[Li et~al.(2022{\natexlab{b}})Li, Wang, He, Li, Wang, Wang, and
  Qiao]{li2022uniformerv2}
Kunchang Li, Yali Wang, Yinan He, Yizhuo Li, Yi~Wang, Limin Wang, and Yu~Qiao.
\newblock {UniFormerV2: Spatiotemporal learning by arming image ViTs with video
  UniFormer}.
\newblock \emph{arXiv preprint arXiv:2211.09552}, 2022{\natexlab{b}}.

\bibitem[Li et~al.(2018)Li, Li, and Vasconcelos]{li2018diving48}
Yingwei Li, Yi~Li, and Nuno Vasconcelos.
\newblock {Resound: Towards action recognition without representation bias}.
\newblock In \emph{ECCV}, 2018.

\bibitem[Lin et~al.(2022)Lin, Geng, Zhang, Gao, de~Melo, Wang, Dai, Qiao, and
  Li]{lin2022frozen}
Ziyi Lin, Shijie Geng, Renrui Zhang, Peng Gao, Gerard de~Melo, Xiaogang Wang,
  Jifeng Dai, Yu~Qiao, and Hongsheng Li.
\newblock {Frozen clip models are efficient video learners}.
\newblock In \emph{ECCV}, 2022.

\bibitem[Liu et~al.(2022)Liu, Bai, and Bai]{liu2022empirical}
Xiaolong Liu, Song Bai, and Xiang Bai.
\newblock {An empirical study of end-to-end temporal action detection}.
\newblock In \emph{CVPR}, 2022.

\bibitem[Monfort et~al.(2019)Monfort, Andonian, Zhou, Ramakrishnan, Bargal,
  Yan, Brown, Fan, Gutfruend, Vondrick, et~al.]{mit}
Mathew Monfort, Alex Andonian, Bolei Zhou, Kandan Ramakrishnan, Sarah~Adel
  Bargal, Tom Yan, Lisa Brown, Quanfu Fan, Dan Gutfruend, Carl Vondrick, et~al.
\newblock {Moments in Time dataset: One million videos for event
  understanding}.
\newblock \emph{IEEE TPAMI}, pp.\  1--8, 2019.

\bibitem[OpenAI(2022)]{gpt4}
OpenAI.
\newblock {GPT-4 Technical Report}.
\newblock \emph{https://cdn.openai.com/papers/gpt-4.pdf}, 2022.

\bibitem[Oquab et~al.(2023)Oquab, Darcet, Moutakanni, Vo, Szafraniec, Khalidov,
  Fernandez, Haziza, Massa, El-Nouby, et~al.]{oquab2023dinov2}
Maxime Oquab, Timoth{\'e}e Darcet, Th{\'e}o Moutakanni, Huy Vo, Marc
  Szafraniec, Vasil Khalidov, Pierre Fernandez, Daniel Haziza, Francisco Massa,
  Alaaeldin El-Nouby, et~al.
\newblock {Dinov2: Learning robust visual features without supervision}.
\newblock In \emph{TMLR}, 2023.

\bibitem[Patraucean et~al.(2024)Patraucean, Smaira, Gupta, Recasens, Markeeva,
  Banarse, Koppula, Malinowski, Yang, Doersch, et~al.]{benchmark-deepmind}
Viorica Patraucean, Lucas Smaira, Ankush Gupta, Adria Recasens, Larisa
  Markeeva, Dylan Banarse, Skanda Koppula, Mateusz Malinowski, Yi~Yang, Carl
  Doersch, et~al.
\newblock {Perception Test: A diagnostic benchmark for multimodal video
  models}.
\newblock In \emph{NeurIPS}, 2024.

\bibitem[Piergiovanni et~al.(2023)Piergiovanni, Kuo, and
  Angelova]{piergiovanni2023rethinking}
AJ~Piergiovanni, Weicheng Kuo, and Anelia Angelova.
\newblock {Rethinking video ViTs: Sparse video tubes for joint image and video
  learning}.
\newblock In \emph{CVPR}, 2023.

\bibitem[Radford et~al.(2021)Radford, Kim, Hallacy, Ramesh, Goh, Agarwal,
  Sastry, Askell, Mishkin, Clark, et~al.]{radford2021clip}
Alec Radford, Jong~Wook Kim, Chris Hallacy, Aditya Ramesh, Gabriel Goh,
  Sandhini Agarwal, Girish Sastry, Amanda Askell, Pamela Mishkin, Jack Clark,
  et~al.
\newblock {Learning transferable visual models from natural language
  supervision}.
\newblock In \emph{ICML}, 2021.

\bibitem[Rajasegaran et~al.(2023)Rajasegaran, Pavlakos, Kanazawa,
  Feichtenhofer, and Malik]{rajasegaran2023benefits}
Jathushan Rajasegaran, Georgios Pavlakos, Angjoo Kanazawa, Christoph
  Feichtenhofer, and Jitendra Malik.
\newblock {On the benefits of 3D pose and tracking for human action
  recognition}.
\newblock In \emph{CVPR}, 2023.

\bibitem[Ramesh et~al.(2021)Ramesh, Pavlov, Goh, Gray, Voss, Radford, Chen, and
  Sutskever]{ramesh2021zero}
Aditya Ramesh, Mikhail Pavlov, Gabriel Goh, Scott Gray, Chelsea Voss, Alec
  Radford, Mark Chen, and Ilya Sutskever.
\newblock {Zero-shot text-to-image generation}.
\newblock In \emph{ICML}, 2021.

\bibitem[Ren et~al.(2015)Ren, He, Girshick, and Sun]{ren2015faster}
Shaoqing Ren, Kaiming He, Ross Girshick, and Jian Sun.
\newblock {Faster R-CNN: Towards real-time object detection with region
  proposal networks}.
\newblock In \emph{NeurIPS}, 2015.

\bibitem[Roberts et~al.(2022)Roberts, Chung, Levskaya, Mishra, Bradbury, Andor,
  Narang, Lester, Gaffney, Mohiuddin, et~al.]{roberts2022t5x}
Adam Roberts, Hyung~Won Chung, Anselm Levskaya, Gaurav Mishra, James Bradbury,
  Daniel Andor, Sharan Narang, Brian Lester, Colin Gaffney, Afroz Mohiuddin,
  et~al.
\newblock {Scaling up models and data with $\texttt{t5x}$ and
  $\texttt{seqio}$}.
\newblock \emph{arXiv preprint arXiv:2203.17189}, 2022.

\bibitem[Sevilla-Lara et~al.(2021)Sevilla-Lara, Zha, Yan, Goswami, Feiszli, and
  Torresani]{sevilla2021only}
Laura Sevilla-Lara, Shengxin Zha, Zhicheng Yan, Vedanuj Goswami, Matt Feiszli,
  and Lorenzo Torresani.
\newblock {Only time can tell: Discovering temporal data for temporal
  modeling}.
\newblock In \emph{WACV}, 2021.

\bibitem[Sigurdsson et~al.(2016)Sigurdsson, Varol, Wang, Farhadi, Laptev, and
  Gupta]{sigurdsson2016charades}
Gunnar~A Sigurdsson, G{\"u}l Varol, Xiaolong Wang, Ali Farhadi, Ivan Laptev,
  and Abhinav Gupta.
\newblock {Hollywood in Homes: Crowdsourcing data collection for activity
  understanding}.
\newblock In \emph{ECCV}, 2016.

\bibitem[Singh et~al.(2022)Singh, Hu, Goswami, Couairon, Galuba, Rohrbach, and
  Kiela]{singh2022flava}
Amanpreet Singh, Ronghang Hu, Vedanuj Goswami, Guillaume Couairon, Wojciech
  Galuba, Marcus Rohrbach, and Douwe Kiela.
\newblock {FLAVA: A foundational language and vision alignment model}.
\newblock In \emph{CVPR}, 2022.

\bibitem[Tong et~al.(2022)Tong, Song, Wang, and Wang]{tong2022videomae}
Zhan Tong, Yibing Song, Jue Wang, and Limin Wang.
\newblock {VideoMAE: Masked autoencoders are data-efficient learners for
  self-supervised video pre-training}.
\newblock \emph{arXiv preprint arXiv:2203.12602}, 2022.

\bibitem[Vaswani et~al.(2017)Vaswani, Shazeer, Parmar, Uszkoreit, Jones, Gomez,
  Kaiser, and Polosukhin]{vaswani2017attention}
Ashish Vaswani, Noam Shazeer, Niki Parmar, Jacob Uszkoreit, Llion Jones,
  Aidan~N. Gomez, Lukasz Kaiser, and Illia Polosukhin.
\newblock Attention is all you need.
\newblock \emph{NeurIPS}, 2017.

\bibitem[Wang et~al.(2018{\natexlab{a}})Wang, Singh, Michael, Hill, Levy, and
  Bowman]{benchmark-glue}
Alex Wang, Amanpreet Singh, Julian Michael, Felix Hill, Omer Levy, and Samuel~R
  Bowman.
\newblock {GLUE: A multi-task benchmark and analysis platform for natural
  language understanding}.
\newblock \emph{arXiv preprint arXiv:1804.07461}, 2018{\natexlab{a}}.

\bibitem[Wang et~al.(2019)Wang, Pruksachatkun, Nangia, Singh, Michael, Hill,
  Levy, and Bowman]{benchmark-superglue}
Alex Wang, Yada Pruksachatkun, Nikita Nangia, Amanpreet Singh, Julian Michael,
  Felix Hill, Omer Levy, and Samuel Bowman.
\newblock {SuperGLUE: A stickier benchmark for general-purpose language
  understanding systems}.
\newblock \emph{Advances in neural information processing systems}, 32, 2019.

\bibitem[Wang et~al.(2018{\natexlab{b}})Wang, Xiong, Wang, Qiao, Lin, Tang, and
  Van~Gool]{wang2018tsn}
Limin Wang, Yuanjun Xiong, Zhe Wang, Yu~Qiao, Dahua Lin, Xiaoou Tang, and Luc
  Van~Gool.
\newblock {Temporal segment networks for action recognition in videos}.
\newblock \emph{IEEE TPAMI}, 41\penalty0 (11):\penalty0 2740--2755,
  2018{\natexlab{b}}.

\bibitem[Wang et~al.(2023)Wang, Huang, Zhao, Tong, He, Wang, Wang, and
  Qiao]{wang2023videomaev2}
Limin Wang, Bingkun Huang, Zhiyu Zhao, Zhan Tong, Yinan He, Yi~Wang, Yali Wang,
  and Yu~Qiao.
\newblock {VideoMAE v2: Scaling video masked autoencoders with dual masking}.
\newblock In \emph{CVPR}, 2023.

\bibitem[Wang et~al.(2022{\natexlab{a}})Wang, Bao, Dong, Bjorck, Peng, Liu,
  Aggarwal, Mohammed, Singhal, Som, et~al.]{wang2022image}
Wenhui Wang, Hangbo Bao, Li~Dong, Johan Bjorck, Zhiliang Peng, Qiang Liu, Kriti
  Aggarwal, Owais~Khan Mohammed, Saksham Singhal, Subhojit Som, et~al.
\newblock {Image as a foreign language: BEIT pretraining for all vision and
  vision-language tasks}.
\newblock \emph{arXiv preprint arXiv:2208.10442}, 2022{\natexlab{a}}.

\bibitem[Wang et~al.(2021)Wang, Qing, Huang, Feng, Zhang, Jiang, Tang, Gao, and
  Sang]{wang2021proposal}
Xiang Wang, Zhiwu Qing, Ziyuan Huang, Yutong Feng, Shiwei Zhang, Jianwen Jiang,
  Mingqian Tang, Changxin Gao, and Nong Sang.
\newblock {Proposal relation network for temporal action detection}.
\newblock \emph{arXiv preprint arXiv:2106.11812}, 2021.

\bibitem[Wang et~al.(2022{\natexlab{b}})Wang, Li, Li, He, Huang, Zhao, Zhang,
  Xu, Liu, Wang, et~al.]{wang2022internvideo}
Yi~Wang, Kunchang Li, Yizhuo Li, Yinan He, Bingkun Huang, Zhiyu Zhao, Hongjie
  Zhang, Jilan Xu, Yi~Liu, Zun Wang, et~al.
\newblock {InternVideo: General video foundation models via generative and
  discriminative learning}.
\newblock \emph{arXiv preprint arXiv:2212.03191}, 2022{\natexlab{b}}.

\bibitem[Wang et~al.(2024)Wang, Li, Li, Yu, He, Chen, Pei, Zheng, Xu, Wang,
  et~al.]{wang2024internvideo2}
Yi~Wang, Kunchang Li, Xinhao Li, Jiashuo Yu, Yinan He, Guo Chen, Baoqi Pei,
  Rongkun Zheng, Jilan Xu, Zun Wang, et~al.
\newblock {InternVideo2: Scaling video foundation models for multimodal video
  understanding}.
\newblock \emph{arXiv preprint arXiv:2403.15377}, 2024.

\bibitem[Wei et~al.(2022)Wei, Tay, Bommasani, Raffel, Zoph, Borgeaud, Yogatama,
  Bosma, Zhou, Metzler, et~al.]{wei2022emergent}
Jason Wei, Yi~Tay, Rishi Bommasani, Colin Raffel, Barret Zoph, Sebastian
  Borgeaud, Dani Yogatama, Maarten Bosma, Denny Zhou, Donald Metzler, et~al.
\newblock {Emergent abilities of large language models}.
\newblock \emph{arXiv preprint arXiv:2206.07682}, 2022.

\bibitem[Xu et~al.(2020)Xu, Zhao, Rojas, Thabet, and Ghanem]{xu2019gtad}
Mengmeng Xu, Chen Zhao, David~S Rojas, Ali Thabet, and Bernard Ghanem.
\newblock {G-TAD: Sub-graph localization for temporal action detection}.
\newblock In \emph{CVPR}, 2020.

\bibitem[Yan et~al.(2022)Yan, Zhu, Wang, Cao, Zhang, Ghosh, Wu, and
  Yu]{yan2022videococa}
Shen Yan, Tao Zhu, Zirui Wang, Yuan Cao, Mi~Zhang, Soham Ghosh, Yonghui Wu, and
  Jiahui Yu.
\newblock {VideoCoCa: Video-text modeling with zero-shot transfer from
  contrastive captioners}.
\newblock \emph{arXiv preprint arXiv:2212.04979}, 2022.

\bibitem[Yang et~al.(2023)Yang, Zhu, Xie, Zhang, Chen, and Li]{yang2023aim}
Taojiannan Yang, Yi~Zhu, Yusheng Xie, Aston Zhang, Chen Chen, and Mu~Li.
\newblock {AIM: Adapting image models for efficient video action recognition}.
\newblock \emph{arXiv preprint arXiv:2302.03024}, 2023.

\bibitem[Yu et~al.(2022)Yu, Wang, Vasudevan, Yeung, Seyedhosseini, and
  Wu]{yu2022coca}
Jiahui Yu, Zirui Wang, Vijay Vasudevan, Legg Yeung, Mojtaba Seyedhosseini, and
  Yonghui Wu.
\newblock {CoCa: Contrastive captioners are image-text foundation models}.
\newblock \emph{arXiv preprint arXiv:2205.01917}, 2022.

\bibitem[Zhang et~al.(2022)Zhang, Wu, and Li]{zhang2022actionformer}
Chen-Lin Zhang, Jianxin Wu, and Yin Li.
\newblock {Actionformer: Localizing moments of actions with transformers}.
\newblock In \emph{ECCV}, 2022.

\bibitem[Zhao et~al.(2024)Zhao, Gundavarapu, Yuan, Zhou, Yan, Sun, Friedman,
  Qian, Weyand, Zhao, et~al.]{zhao2024videoprism}
Long Zhao, Nitesh~B Gundavarapu, Liangzhe Yuan, Hao Zhou, Shen Yan, Jennifer~J
  Sun, Luke Friedman, Rui Qian, Tobias Weyand, Yue Zhao, et~al.
\newblock {VideoPrism: A foundational visual encoder for video understanding}.
\newblock In \emph{ICML}, 2024.

\end{thebibliography}
}

\end{document}